\documentclass[letterpaper, 10 pt, journal, twoside]{IEEEtran}

\IEEEoverridecommandlockouts

\usepackage{microtype}

\title{Robust Visual Localization in Compute-Constrained Environments by Salient Edge Rendering and Weighted Hamming Similarity
}

\author{Tu-Hoa Pham$^{1}$, Philip Bailey$^2$, Daniel Posada$^2$, Georgios Georgakis$^{1}$, \\ Jorge Enriquez$^3$, Surya Suresh$^4$, Marco Dolci$^1$ and Philip Twu$^1$\thanks{This research was carried out at the Jet Propulsion Laboratory, California Institute of Technology, under a contract with the National Aeronautics and Space Administration (80NM0018D0004). The decision to implement Mars Sample Return will not be finalized until NASA’s completion of the National Environmental Policy Act (NEPA) process. This document is being made available for information purposes only.}\thanks{$^{1}$T.-H. Pham, G. Georgakis, M. Dolci and P. Twu are with the Jet Propulsion Laboratory, California Institute of Technology, 4800 Oak Grove Drive, Pasadena, CA 91109, USA
        {\tt\small firstname.lastname@jpl.nasa.gov}}\thanks{$^{2}$P. Bailey and D. Posada are with Blue Origin, 2001 Edmund Halley Drive, Reston, VA 20191, USA
        {\tt\small pbailey@blueorigin.com}, {\tt\small dposada@blueorigin.com}}\thanks{$^{3}$J. Enriquez is with Amazon, 400 9th Ave N, Seattle, WA 98109, USA
        {\tt\small jorgeeqz@amazon.com}}\thanks{$^{4}$S. Suresh is with Ohio State University, 281 W Lane Ave, Columbus, OH 43210, USA
        {\tt\small suresh.163@buckeyemail.osu.edu}}
        \thanks{Digital Object Identifier (DOI): 10.1109/LRA.2025.3614045}
        \thanks{\copyright 2025 IEEE. All rights reserved, including rights for text and data mining and training of artificial intelligence and similar technologies. Personal use is permitted,
but republication/redistribution requires IEEE permission. See https://www.ieee.org/publications/rights/index.html for more information.}
}

\usepackage[utf8]{inputenc}
\usepackage{graphicx}
\usepackage{amsmath}
\usepackage{mathtools}
\usepackage{amssymb}
\usepackage{textcomp}
\usepackage{graphicx}
\usepackage{subcaption}
\usepackage{color}
\usepackage[dvipsnames]{xcolor}
\usepackage{array}
\usepackage[percent]{overpic}
\usepackage{tikz}
\usetikzlibrary{calc}
\usepackage{balance}
\usepackage{multirow,booktabs}
\usepackage{siunitx}
\usepackage{csquotes}
\usepackage[export]{adjustbox}
\usepackage{nicefrac}
\usepackage{comment}
\usepackage{arydshln}

\usepackage{algpseudocode,algorithm,algorithmicx}

\algrenewcommand\algorithmicrequire{\textbf{Precondition:}}
\algrenewcommand\algorithmicensure{\textbf{Postcondition:}}
\algnewcommand\algorithmicforeach{\textbf{for each}}
\algdef{S}[FOR]{ForEach}[1]{\algorithmicforeach\ #1\ \algorithmicdo}

\newcommand{\MStationFrame}{S}
\newcommand{\MCameraFrame}{C}
\newcommand{\MCameraEstimateFrame}{\widehat{\MCameraFrame}}
\newcommand{\MEndEffectorFrame}{EE}
\newcommand{\MEndEffectorEstimateFrame}{\widehat{\MEndEffectorFrame}}

\newcommand{\MEdgeMarker}{+}
\newcommand{\MNotEdgeMarker}{-}

\newcommand{\MTestElement}{I}
\newcommand{\MTestImage}{\mathbf{\MTestElement}}
\newcommand{\MTestEdgeImage}{\MTestImage^{\MEdgeMarker}}
\newcommand{\MTestNotEdgeImage}{\MTestImage^{\MNotEdgeMarker}}
\newcommand{\MTestSizeCol}{s_{x}}
\newcommand{\MTestSizeRow}{s_{y}}
\newcommand{\MTestSize}{\MTestSizeCol \times \MTestSizeRow}
\newcommand{\MBaselineImage}{\widehat{\mathbf{\MTestElement}}}
\newcommand{\MTemplateElement}{\widehat{T}}
\newcommand{\MTemplateImage}{\mathbf{\MTemplateElement}}
\newcommand{\MTemplateSizeCol}{\widehat{s}_{x}}
\newcommand{\MTemplateSizeRow}{\widehat{s}_{y}}
\newcommand{\MTemplateSize}{\MTemplateSizeCol \times \MTemplateSizeRow}
\newcommand{\MTemplatePixelCountEdge}{c^{\MEdgeMarker}}
\newcommand{\MTemplatePixelCountNotEdge}{c^{\MNotEdgeMarker}}

\newcommand{\MTemplateWeightEdge}{w^{\MEdgeMarker}}
\newcommand{\MTemplateWeightNotEdge}{w^{\MNotEdgeMarker}}
\newcommand{\MTemplateMaskedEdgeImage}{\MTemplateImage^{\MEdgeMarker}}
\newcommand{\MTemplateMaskedNotEdgeImage}{\MTemplateImage^{\MNotEdgeMarker}}
\newcommand{\MTemplateMaskedEdgeImageReversed}{\overline{\MTemplateMaskedEdgeImage}}
\newcommand{\MTemplateMaskedNotEdgeImageReversed}{\overline{\MTemplateMaskedNotEdgeImage}}
\newcommand{\MMaskElement}{\widehat{M}}
\newcommand{\MMaskImage}{\mathbf{\MMaskElement}}

\newcommand{\MScoreElement}{S}
\newcommand{\MScoreMatrix}{\mathbf{\MScoreElement}}

\newcommand{\MTransform}[2]{{}^{#1}\mathbf{T}_{#2}}
\newcommand{\MTemplateCount}{n}
\newcommand{\MTestPixel}{p^{\text{2D}}}

\newcommand{\MBaselinePixel}{\widehat{p}^{\text{2D}}}
\newcommand{\MBaselinePoint}{\widehat{p}^{\text{3D}}}

\newcommand{\MDepthComponent}{\widehat{D}}
\newcommand{\MDepthBuffer}{\mathbf{\MDepthComponent}}

\newcommand{\MPnPMultipleCount}{N^{\text{mult}}}

\newcommand{\todo}[1]{\textcolor{red}{TODO: #1}\PackageWarning{todo}{Remove TODO from final version ("#1")}}

\graphicspath{{figures/}}
\DeclareGraphicsExtensions{.pdf,.jpg,.jpeg,.png}

\newcommand{\Mmakeconopsandresultsfigure}[1]{
\begin{figure}[#1]
\centering
    \begin{tikzpicture}
\node[anchor=south west, inner sep=0] (image) at (0,0) {\includegraphics[width=0.985\columnwidth]{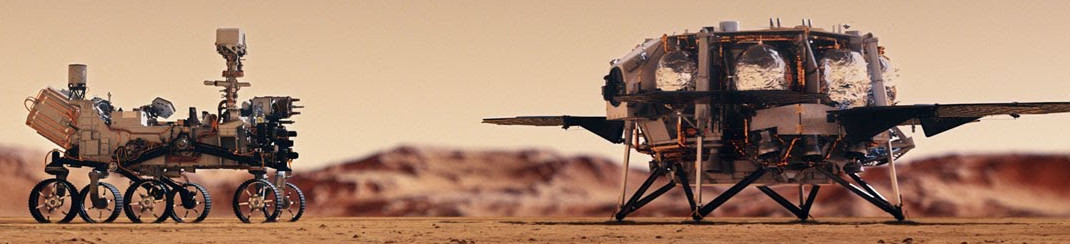}};
      \begin{scope}[x={(image.south east)}, y={(image.north west)}]  

\node[draw, fill=white, rounded corners, font=\bfseries] (lander) at (0.42,0.80) {Lander OS};
        \draw[->, thick] (lander) |- (0.56,0.63);  

\node[draw, fill=white, rounded corners, font=\bfseries] (rover) at (0.42,0.20) {Rover BC};
        \draw[->, thick] (rover) |- (0.277,0.48);  

      \end{scope}
    \end{tikzpicture} \\
    \vspace{0.08cm}
    \begin{tikzpicture}
\node[anchor=south west, inner sep=0] (image) at (0,0) {
             \includegraphics[height=0.35\columnwidth]{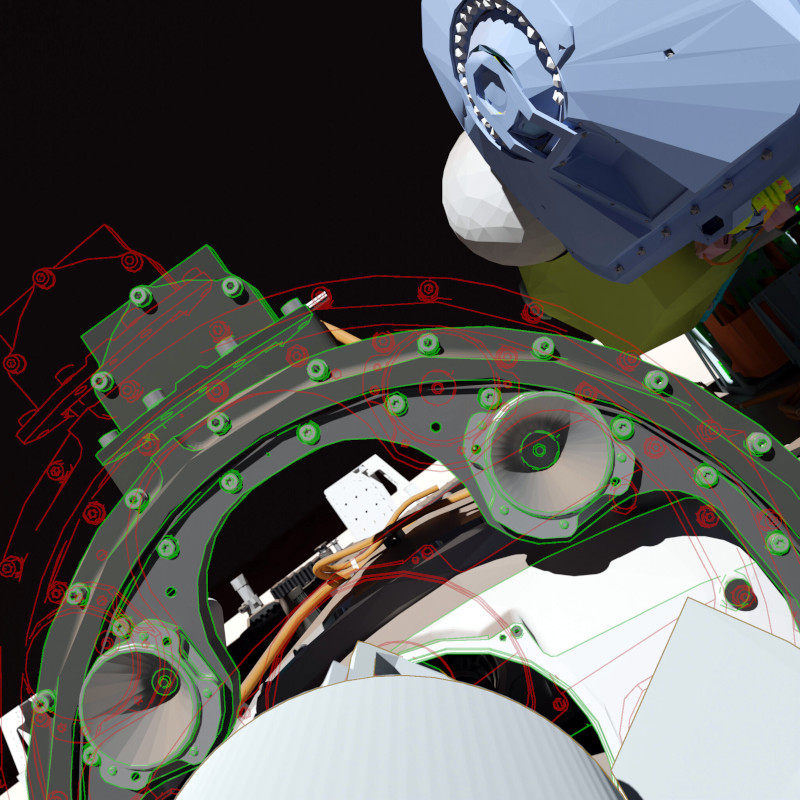}
         };
\node[fill=lightgray, opacity=0.8, text opacity=1, anchor=north east, align=right, inner ysep=1pt] at (image.north east) {
             Synthetic BC
         };
    \end{tikzpicture}
    \begin{tikzpicture}
\node[anchor=south west, inner sep=0] (image) at (0,0) {
             \includegraphics[height=0.35\columnwidth]{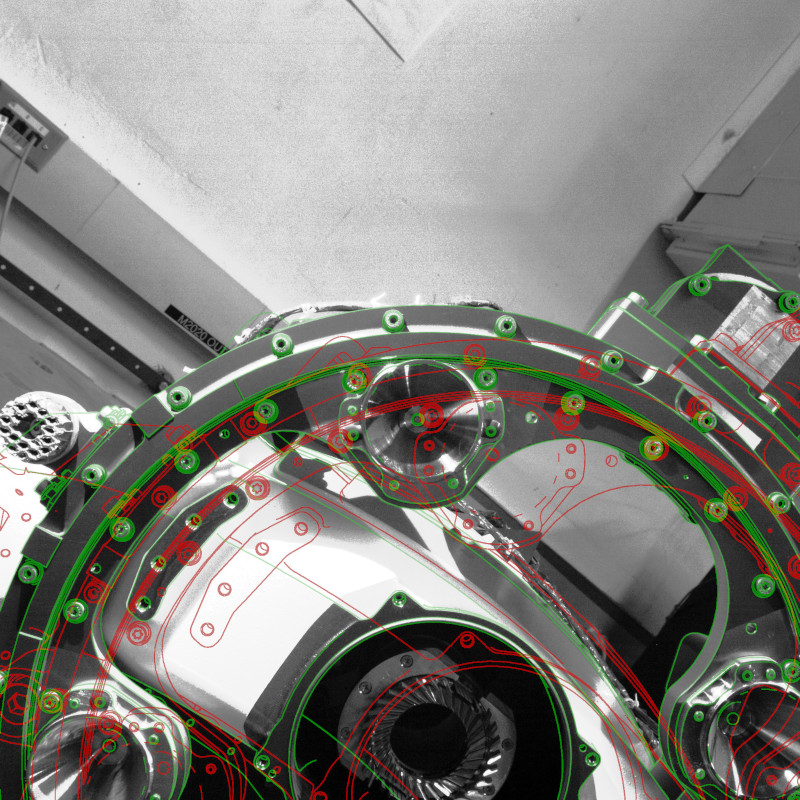}
         };
\node[fill=lightgray, opacity=0.8, text opacity=1, anchor=north east, align=right, inner ysep=1pt] at (image.north east) {
             Testbed BC
         };
    \end{tikzpicture}
    \begin{tikzpicture}
\node[anchor=south west, inner sep=0] (image) at (0,0) {
             \includegraphics[height=0.35\columnwidth]{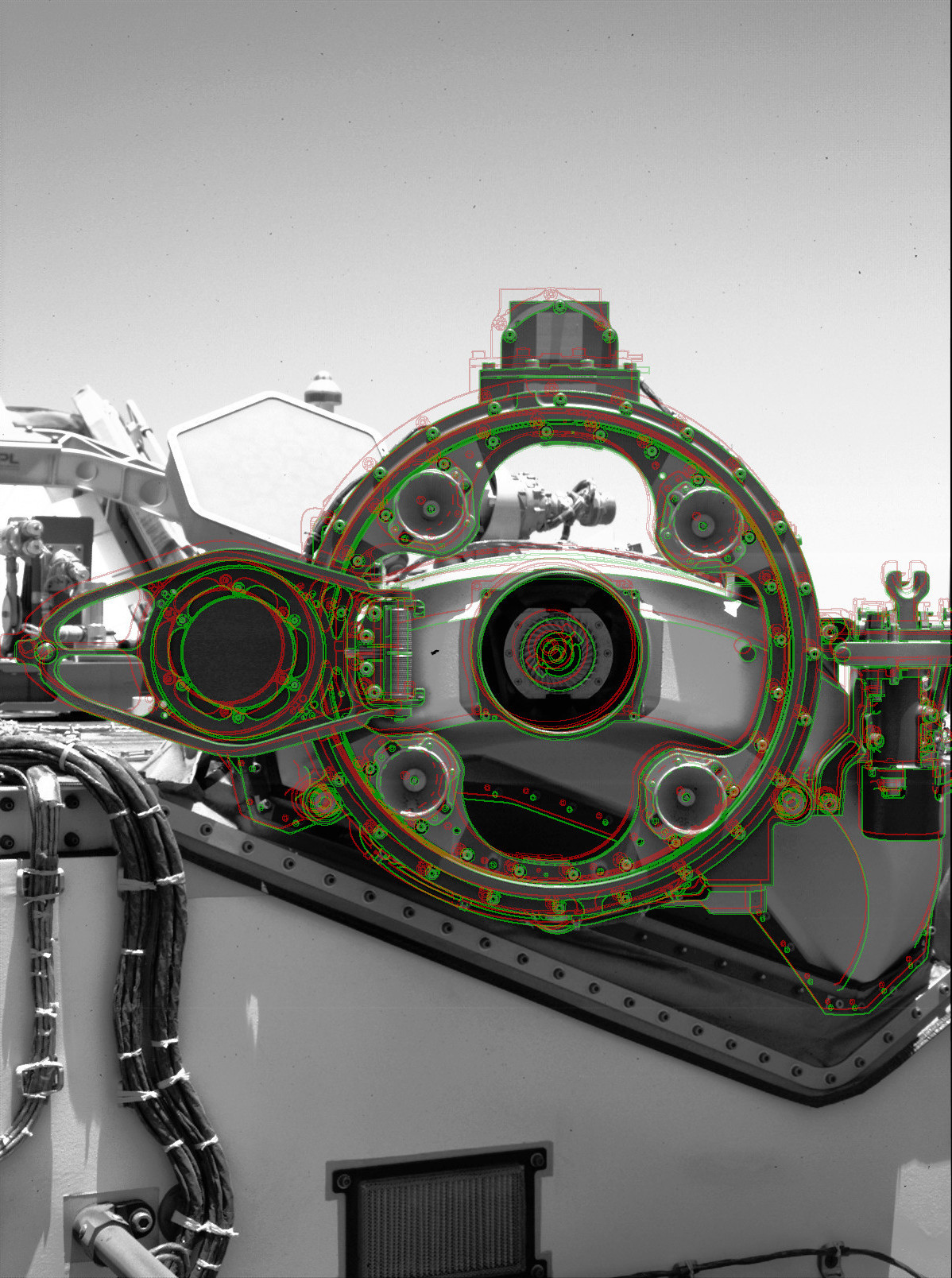}
         };
\node[fill=lightgray, opacity=0.8, text opacity=1, anchor=north east, align=right, inner xsep=1pt, inner ysep=1pt] at (image.north east) {
             Mars far BC
         };
    \end{tikzpicture} \\
    \vspace{0.08cm}
    \begin{tikzpicture}
\node[anchor=south west, inner sep=0] (image) at (0,0) {
             \includegraphics[height=0.35\columnwidth]{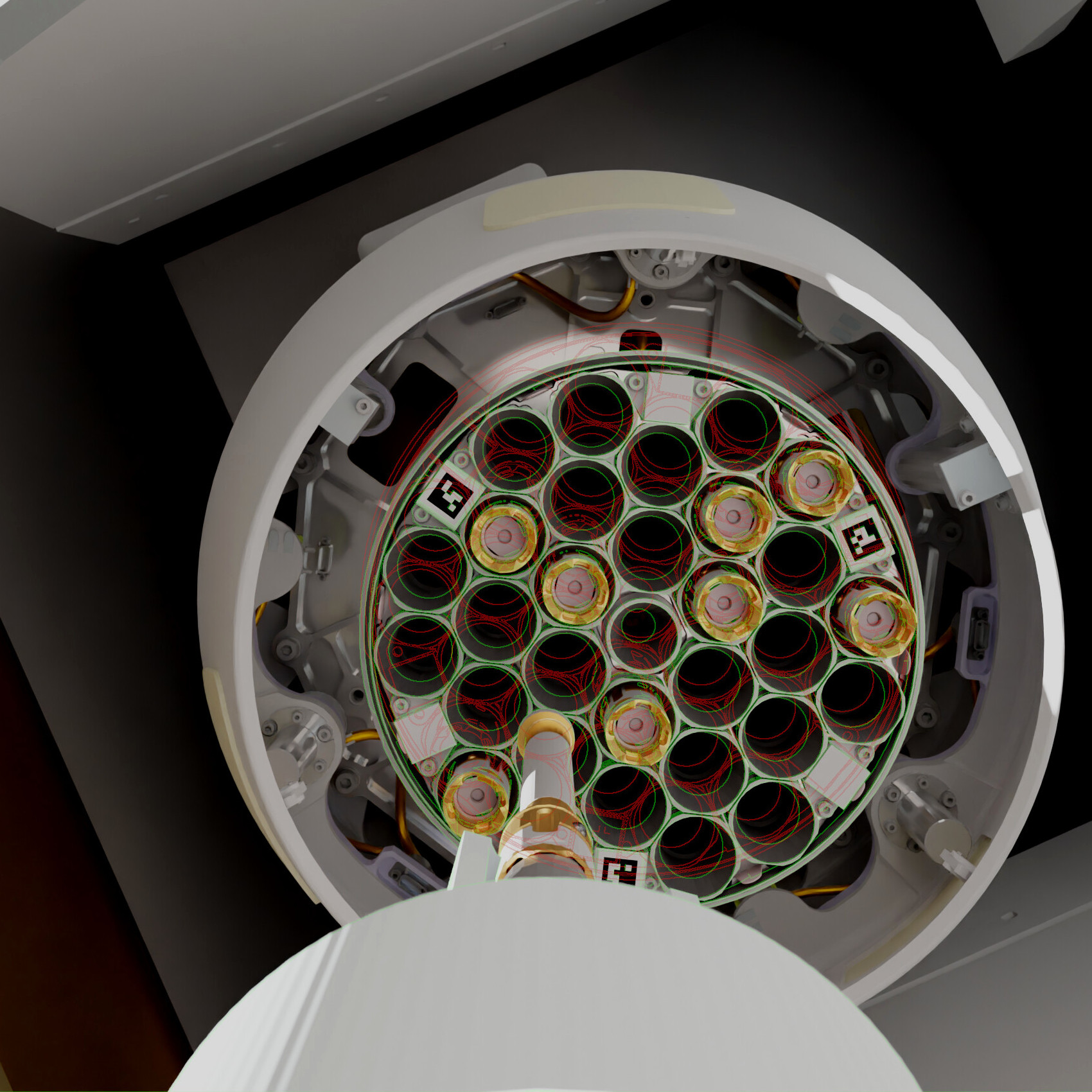}
         };
\node[fill=lightgray, opacity=0.8, text opacity=1, anchor=north east, align=right, inner ysep=1pt] at (image.north east) {
             Synthetic OS
         };
    \end{tikzpicture}
    \begin{tikzpicture}
\node[anchor=south west, inner sep=0] (image) at (0,0) {
             \includegraphics[height=0.35\columnwidth]{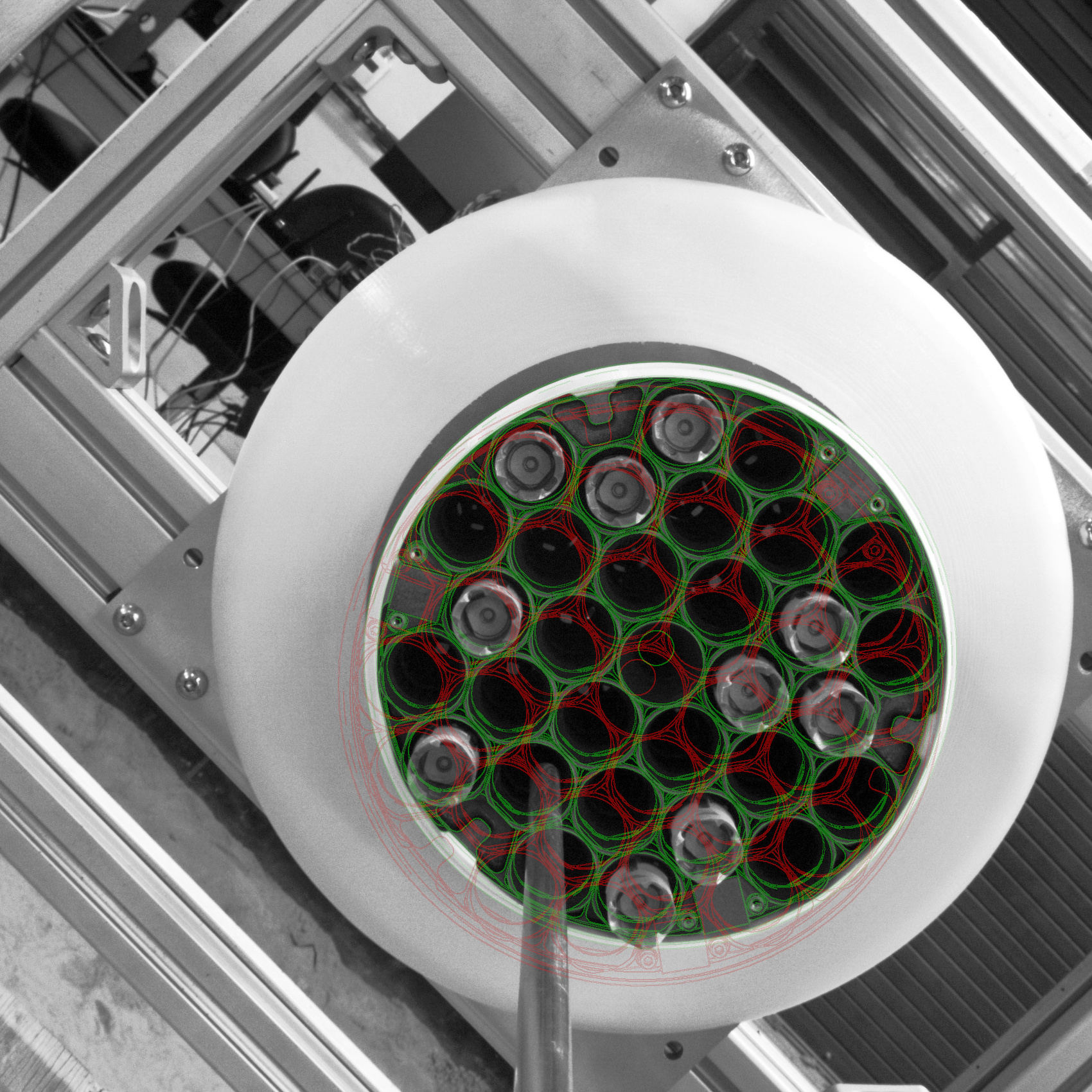}
         };
\node[fill=lightgray, opacity=0.8, text opacity=1, anchor=north east, align=right, inner ysep=1pt] at (image.north east) {
             Testbed OS
         };
    \end{tikzpicture}
    \begin{tikzpicture}
\node[anchor=south west, inner sep=0] (image) at (0,0) {
             \includegraphics[height=0.35\columnwidth]{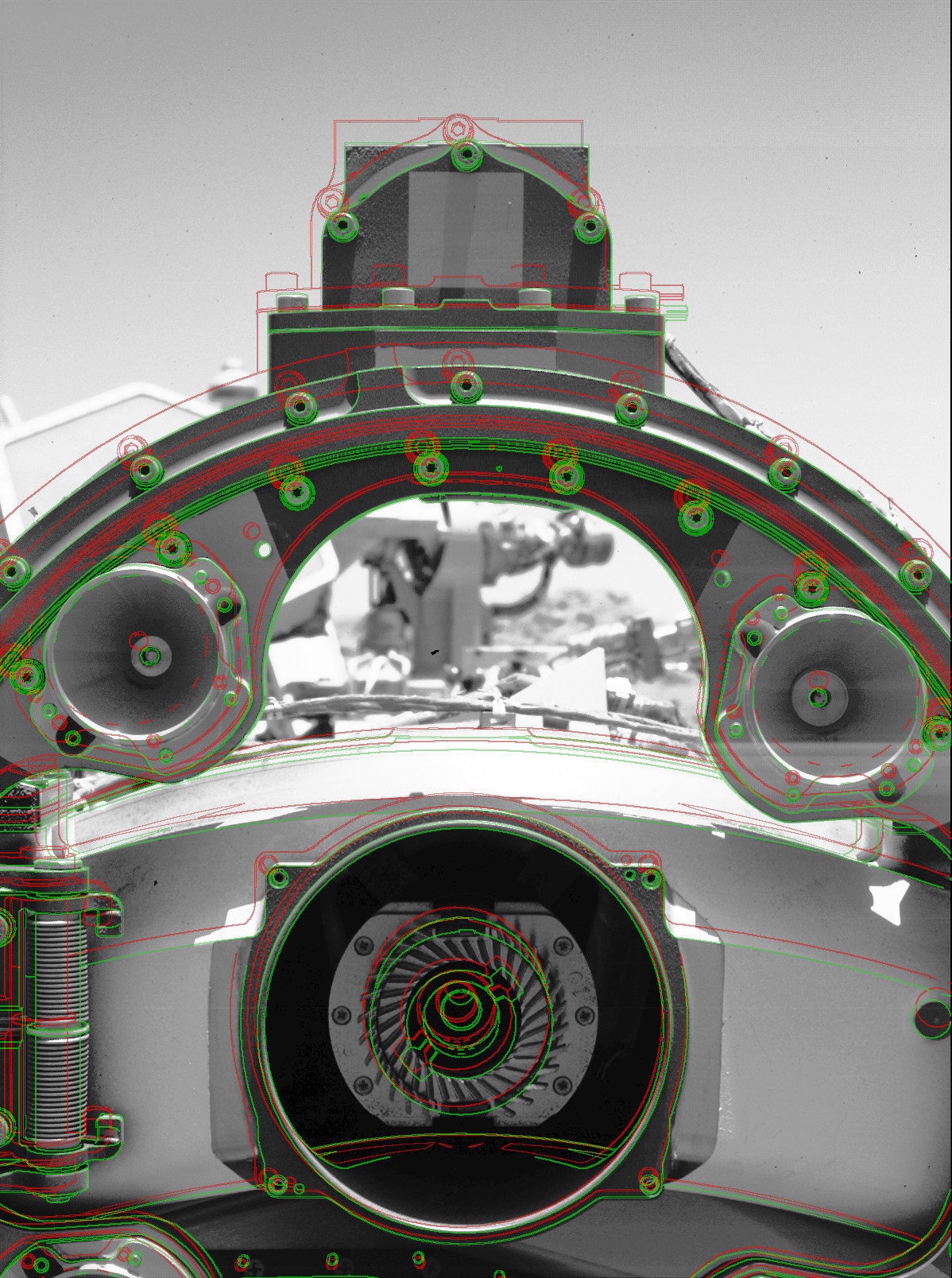}
         };
\node[fill=lightgray, opacity=0.8, text opacity=1, anchor=north east, align=right, inner xsep=1pt, inner ysep=1pt] at (image.north east) {
             Mars near BC
         };
    \end{tikzpicture}
    \caption{
        Top: concept view of the Perseverance rover meeting the Sample Return Lander (courtesy of NASA)
        to transfer samples from bit carousel (BC) to orbiting sample (OS) canister.
        Middle and bottom: seed (red) and pose from visual localization (green)
        on synthetic, testbed and Mars data.
    }
    \label{fig:conops_and_results}
\end{figure}
}

\newcommand{\Mmakealgorithmfigure}[1]{
\begin{figure}[#1]
    \smallskip
    \smallskip
    \centering
    \includegraphics[height=0.32\columnwidth]{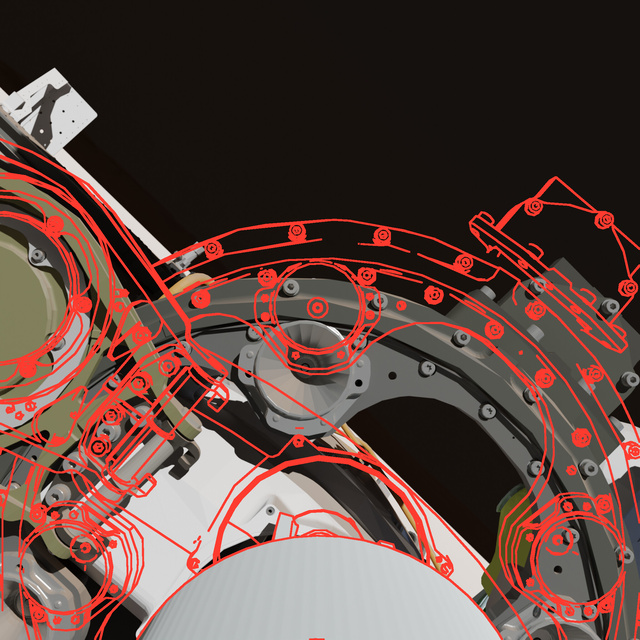}
    \includegraphics[height=0.32\columnwidth]{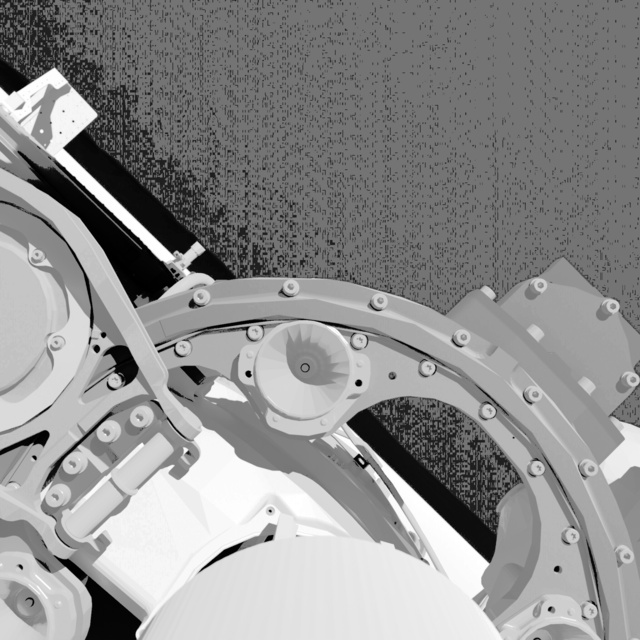}
    \includegraphics[height=0.32\columnwidth]{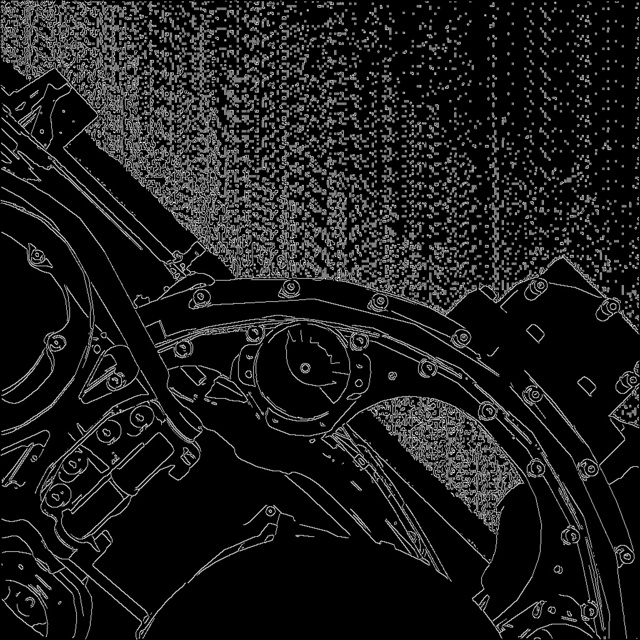}
    \includegraphics[height=0.323\columnwidth]{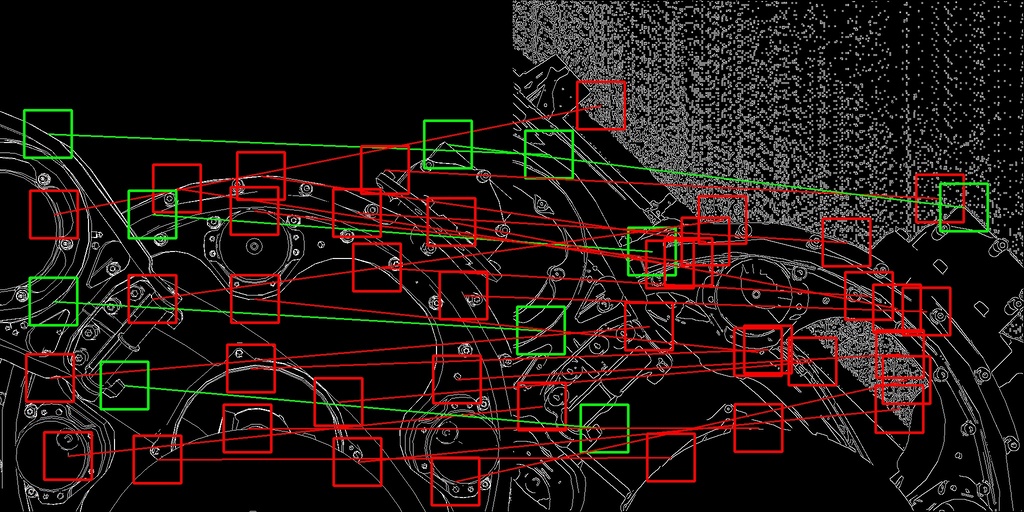}
    \includegraphics[height=0.323\columnwidth]{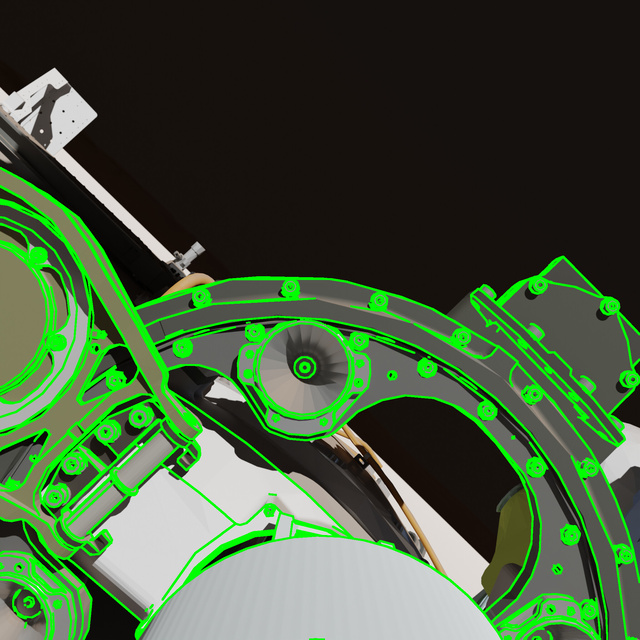}
    \caption{
        Pose estimation pipeline.
        Top: input image with seed pose overlaid in red, after histogram equalization, and after Canny edge detection (notice the background noise).
        Bottom: template matching using Weighted Hamming Similarity against rendered salient edges, and estimated pose in green.
    }
    \label{fig:tsm}
\end{figure}
}

\newcommand{\Mmakedatasetfigure}[1]{
\begin{figure}[#1]
    \smallskip
    \smallskip
    \centering
    \begin{subfigure}[t]{1.0\columnwidth}
        \centering
\begin{tikzpicture}
\node[anchor=south west, inner sep=0] (image) at (0,0) {\includegraphics[height=0.33\columnwidth]{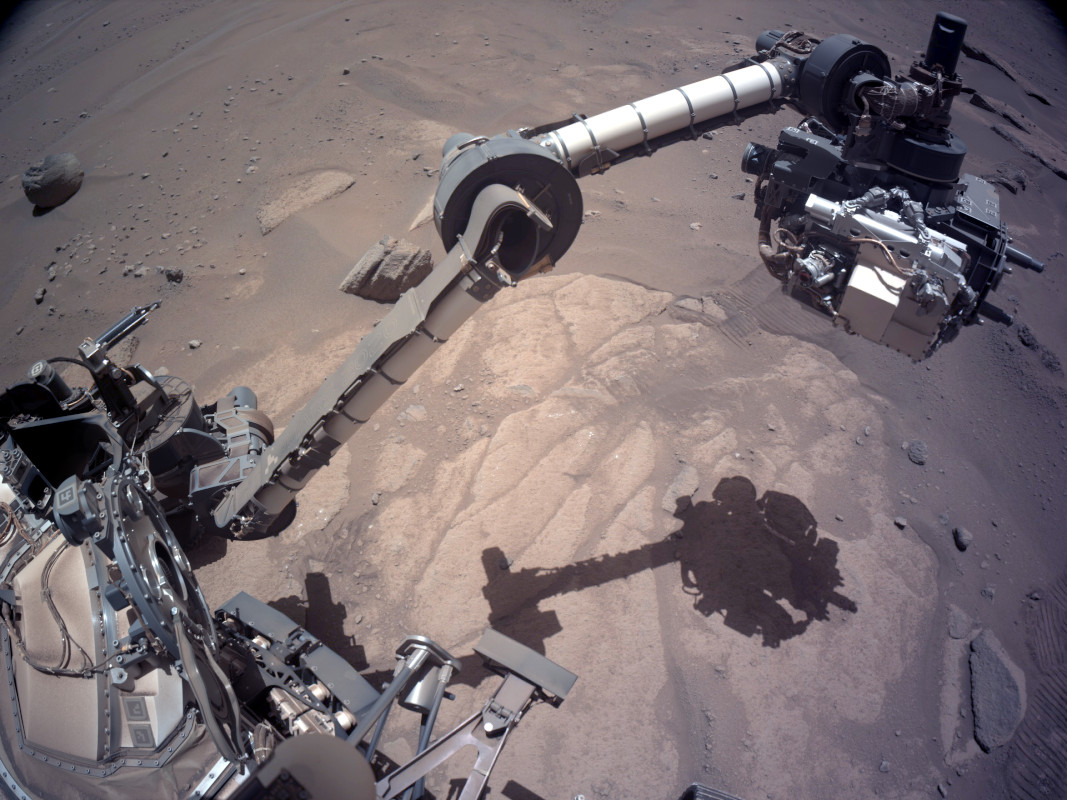}};
          \begin{scope}[x={(image.south east)}, y={(image.north west)}]  

\node[draw, fill=white, opacity=0.8, text opacity=1, rounded corners, font=\scriptsize\bfseries] (lander) at (0.25,0.90) {Rover camera};
            \draw[->, thick, white] (lander) |- (0.72,0.65);  

\node[draw, fill=white, opacity=0.8, text opacity=1, rounded corners, font=\scriptsize\bfseries] (rover) at (0.800,0.29) {Rover BC};
            \draw[->, thick, white] (rover) -- (0.20,0.29);  

\draw[->, thick, white, dashed] (0.73,0.63) -- (0.18,0.31);

          \end{scope}
        \end{tikzpicture}
\begin{tikzpicture}
\node[anchor=south west, inner sep=0] (image) at (0,0) {
                \includegraphics[height=0.33\columnwidth]{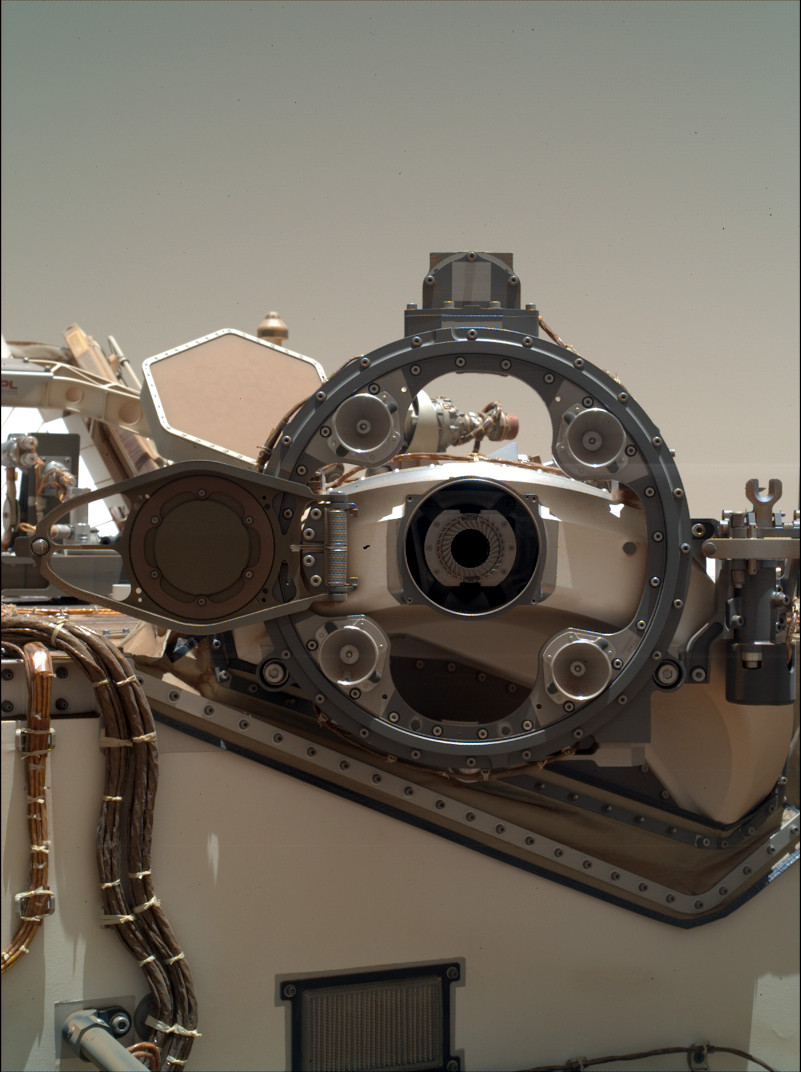}
            };
\node[fill=white, opacity=0.8, text opacity=1, anchor=north east, align=right, inner sep=1pt, font=\scriptsize\bfseries] at (image.north east) {
                Mars BC, far
            };
        \end{tikzpicture}
\begin{tikzpicture}
\node[anchor=south west, inner sep=0] (image) at (0,0) {
                \includegraphics[height=0.33\columnwidth]{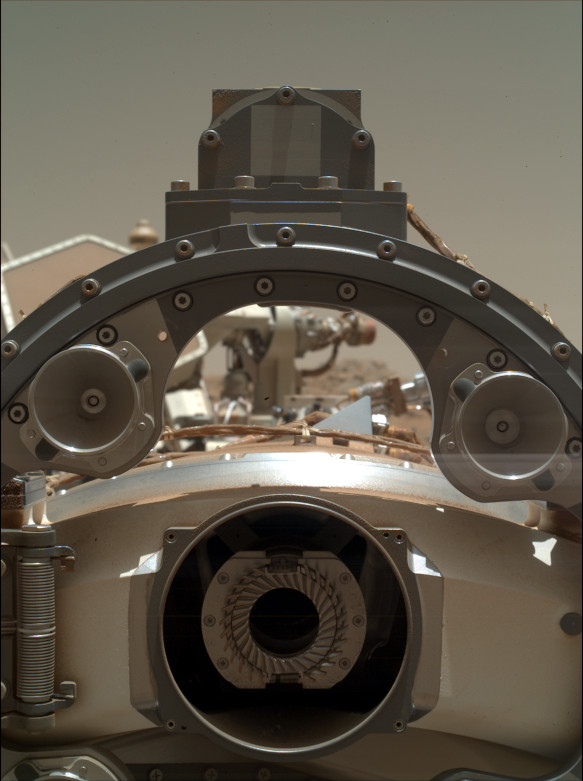}
            };
\node[fill=white, opacity=0.8, text opacity=1, anchor=north east, align=right, inner sep=1pt, font=\scriptsize\bfseries] at (image.north east) {
                Mars BC, near
            };
        \end{tikzpicture}
    \end{subfigure} \\
    \vspace{0.08cm}
    \begin{subfigure}[t]{1.0\columnwidth}
        \centering
\begin{tikzpicture}
\node[anchor=south west, inner sep=0] (image) at (0,0) {\includegraphics[height=0.28\columnwidth]{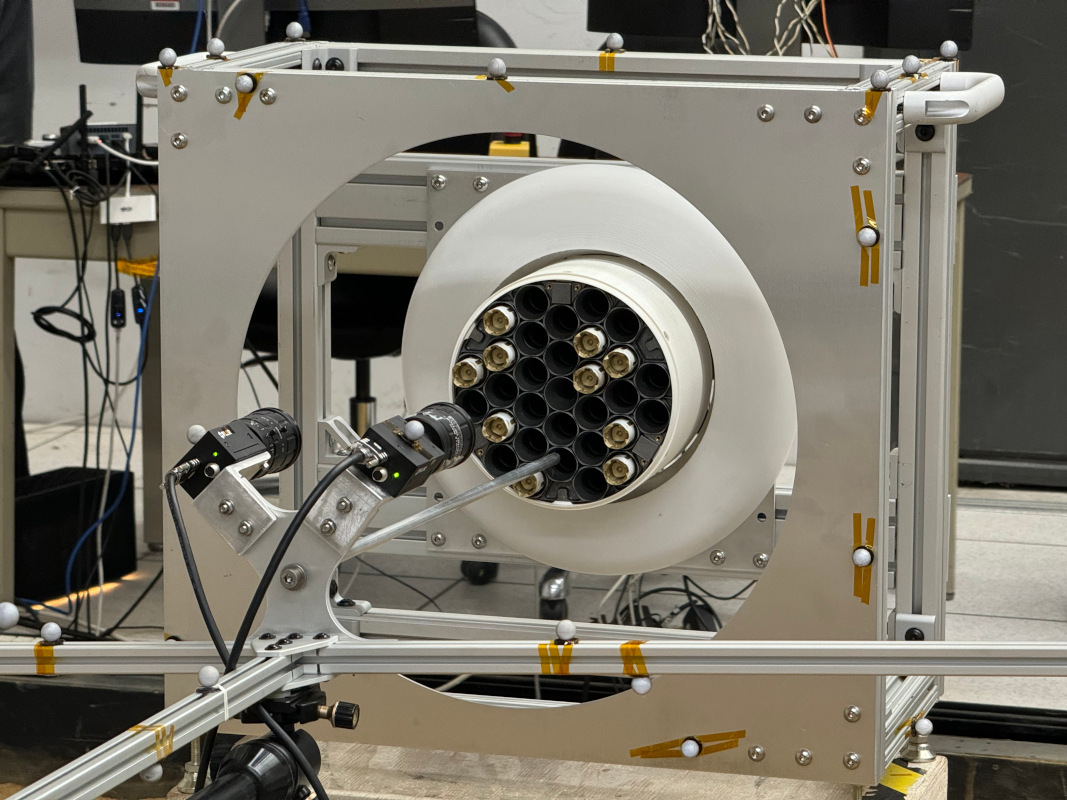}};
          \begin{scope}[x={(image.south east)}, y={(image.north west)}]  \node[draw, fill=white, opacity=0.8, text opacity=1, rounded corners, font=\scriptsize\bfseries] (testbed) at (0.32,0.10) {Testbed cameras};
\coordinate (r_target) at (0.37, 0.37);
\coordinate (r_intermediate) at (0.37, 0.185); \draw[->, thick, white] (r_intermediate) -- (r_target);
\coordinate (l_target) at (0.20, 0.37);
\coordinate (l_intermediate) at (0.20, 0.185); \draw[->, thick, white] (l_intermediate) -- (l_target);
          \end{scope}
        \end{tikzpicture}
\begin{tikzpicture}
\node[anchor=south west, inner sep=0] (image) at (0,0) {
                \includegraphics[height=0.28\columnwidth]{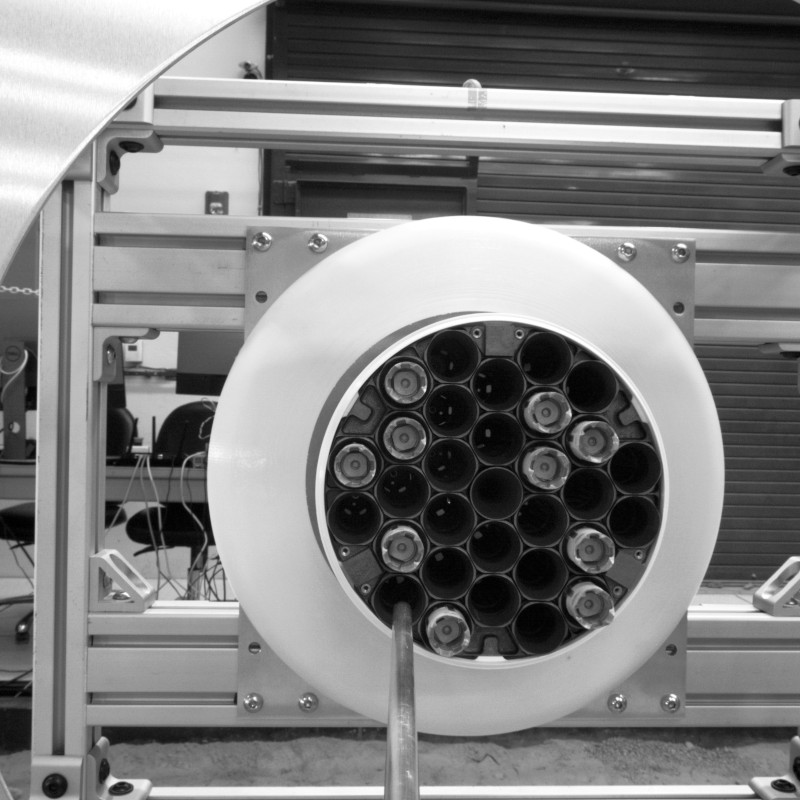}
            };
\node[fill=white, opacity=0.8, text opacity=1, anchor=north east, align=right, inner sep=1pt, font=\scriptsize\bfseries] at (image.north east) {
                Testbed OS, left
            };
        \end{tikzpicture}
\begin{tikzpicture}
\node[anchor=south west, inner sep=0] (image) at (0,0) {
                \includegraphics[height=0.28\columnwidth]{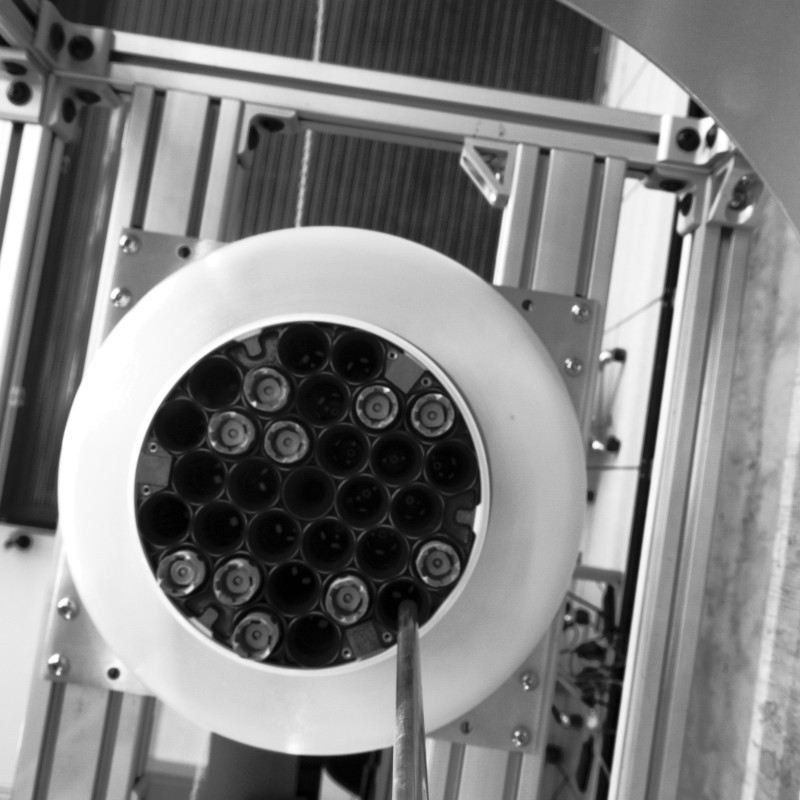}
            };
\node[fill=white, opacity=0.8, text opacity=1, anchor=north east, align=right, inner sep=1pt, font=\scriptsize\bfseries] at (image.north east) {
                Testbed OS, right
            };
        \end{tikzpicture}
\end{subfigure} \\
    \vspace{0.08cm}
    \begin{subfigure}[t]{1.0\columnwidth}
        \centering
\begin{tikzpicture}
\node[anchor=south west, inner sep=0] (image) at (0,0) {\includegraphics[height=0.28\columnwidth]{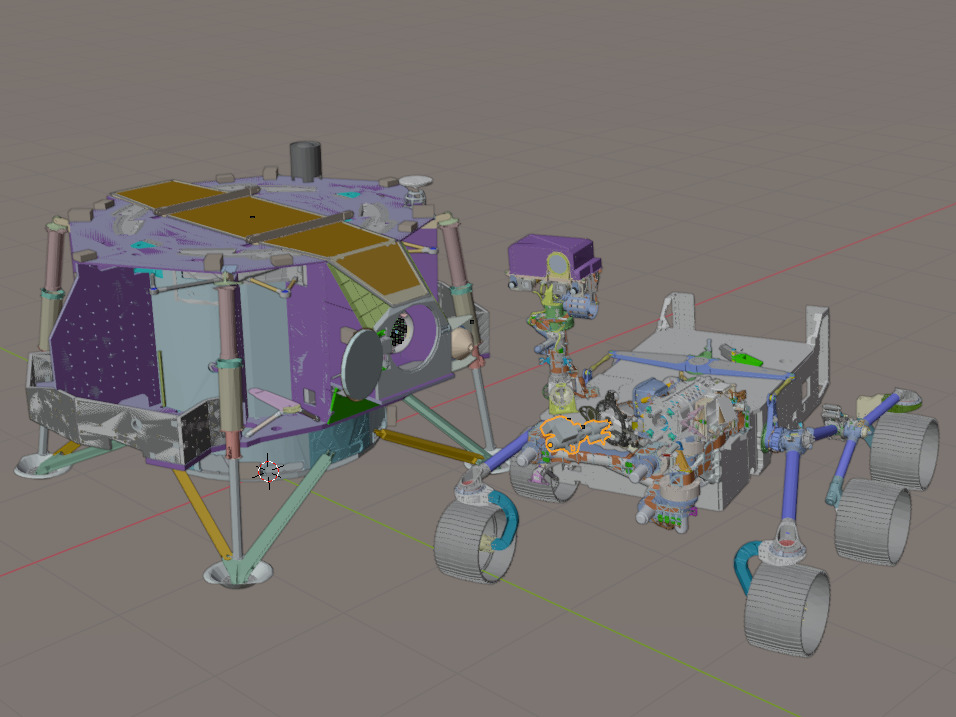}};
          \begin{scope}[x={(image.south east)}, y={(image.north west)}]  \node[draw, fill=white, opacity=0.8, text opacity=1, rounded corners, font=\scriptsize\bfseries] (os) at (0.25,0.90) {Lander OS};
            \draw[->, thick, white] (os) |- (0.39,0.54);  \node[draw, fill=white, opacity=0.8, text opacity=1, rounded corners, font=\scriptsize\bfseries] (bc) at (0.74,0.80) {Rover BC};
            \draw[->, thick, white] (bc) |- (0.67,0.41);  \node[draw, fill=white, opacity=0.8, text opacity=1, rounded corners, font=\scriptsize\bfseries] (lander_cameras) at (0.52,0.10) {Lander cameras};
\coordinate (target) at (0.59, 0.35);
\coordinate (intermediate) at (0.59, 0.185); \draw[->, thick, white] (intermediate) -- (target);
          \end{scope}
        \end{tikzpicture}
\begin{tikzpicture}
\node[anchor=south west, inner sep=0] (image) at (0,0) {
                \includegraphics[height=0.28\columnwidth]{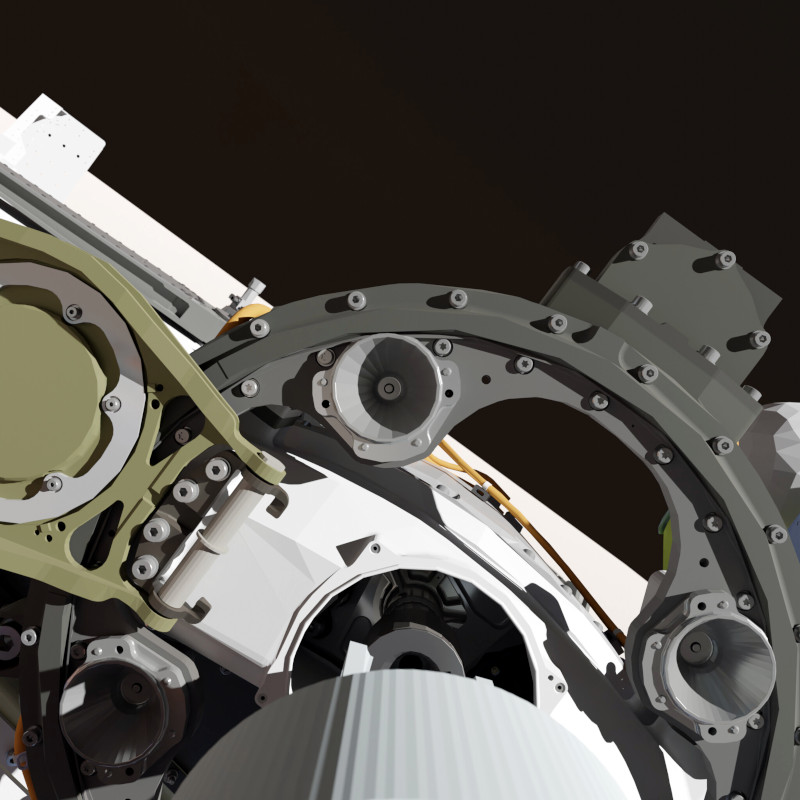}
            };
\node[fill=white, opacity=0.8, text opacity=1, anchor=north east, align=right, inner sep=1pt, font=\scriptsize\bfseries] at (image.north east) {
                Synthetic BC, left
            };
        \end{tikzpicture}
\begin{tikzpicture}
\node[anchor=south west, inner sep=0] (image) at (0,0) {
                \includegraphics[height=0.28\columnwidth]{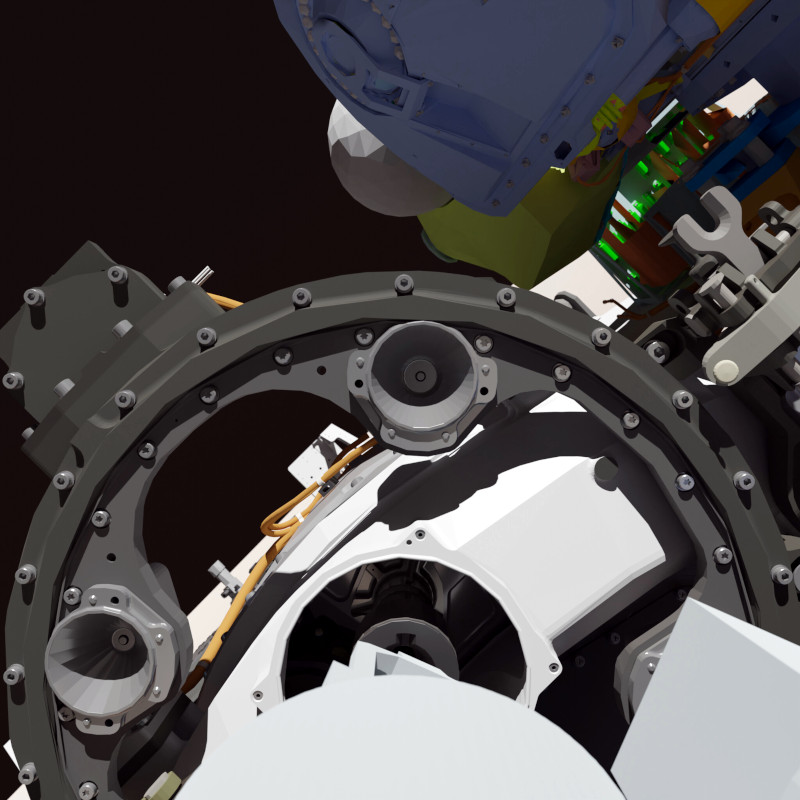}
            };
\node[fill=white, opacity=0.8, text opacity=1, anchor=north east, align=right, inner sep=1pt, font=\scriptsize\bfseries] at (image.north east) {
                Synthetic BC, right
            };
        \end{tikzpicture}
\end{subfigure}
\caption{Three datasets for evaluation: Mars 2020 rover pointing its arm camera at the BC, far and near shot (top), lander OS testbed, left and right images (middle), virtual scene with lander cameras on rover BC, left and right renders (bottom).}
    \label{fig:dataset_full}
\end{figure}
}

\newcommand{\Mmakeresultsfigurewithloftr}[0]{
\begin{figure*}[ht]
    \smallskip
    \smallskip
    \centering
    \includegraphics[width=0.95\textwidth]{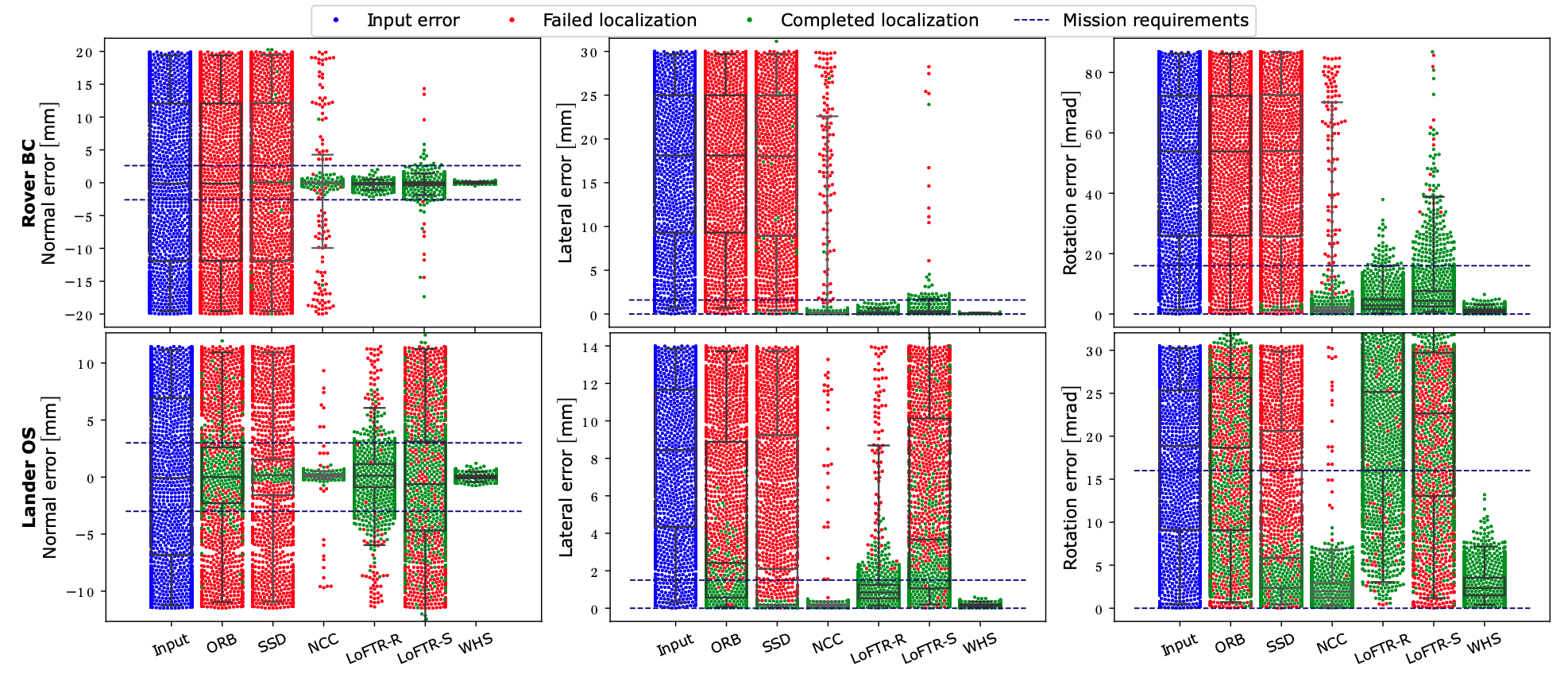}
\caption{Error distribution per method and dimension on synthetic data for rover BC (top) and lander OS (bottom).}
    \label{fig:results_sim_with_loftr}
\end{figure*}
}

\newcommand{\Mmakeextensionfigurewithtextboxes}[1]{
\begin{figure}[#1]
    \smallskip
    \smallskip
    \centering
    \begin{subfigure}[t]{1.0\columnwidth}
        \centering
       \begin{tikzpicture}
\node[anchor=south west, inner sep=0] (image) at (0,0) {
                \includegraphics[width=0.48\columnwidth]{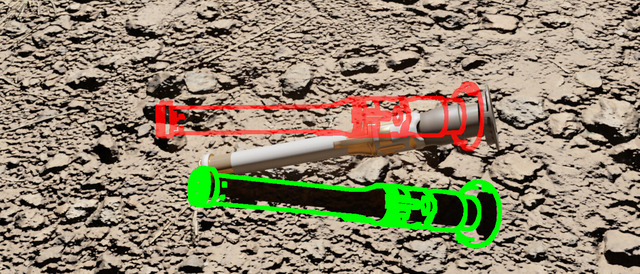}
            };
\node[fill=lightgray, opacity=0.8, text opacity=1, anchor=north east, align=right, inner sep=1pt] at (image.north east) {
                Tube failure \textbf{\textcolor{red}{seed}} and \textbf{\textcolor{green}{end}}
            };
        \end{tikzpicture}
       \begin{tikzpicture}
\node[anchor=south west, inner sep=0] (image) at (0,0) {
                \includegraphics[width=0.48\columnwidth]{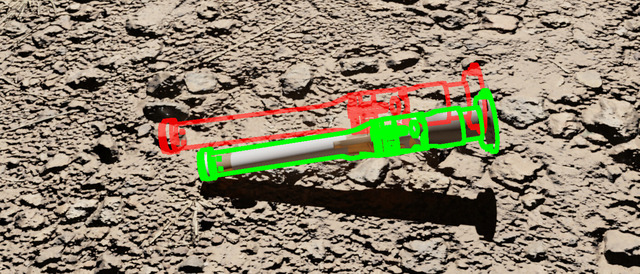}
            };
\node[fill=lightgray, opacity=0.8, text opacity=1, anchor=north east, align=right, inner sep=1pt] at (image.north east) {
                Tube success \textbf{\textcolor{red}{seed}} and \textbf{\textcolor{green}{end}}
            };
        \end{tikzpicture}
    \end{subfigure} \\
    \vspace{0.08cm}
    \begin{subfigure}[t]{1.0\columnwidth}
        \centering
       \begin{tikzpicture}
\node[anchor=south west, inner sep=0] (image) at (0,0) {
                \includegraphics[width=0.48\columnwidth]{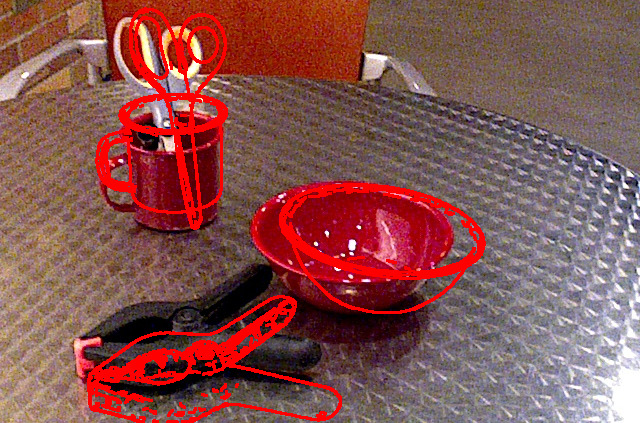}
            };
\node[fill=lightgray, opacity=0.8, text opacity=1, anchor=north east, align=right, inner sep=1pt] at (image.north east) {
                YCB dataset \textbf{\textcolor{red}{seed}}
            };
        \end{tikzpicture}
       \begin{tikzpicture}
\node[anchor=south west, inner sep=0] (image) at (0,0) {
                \includegraphics[width=0.48\columnwidth]{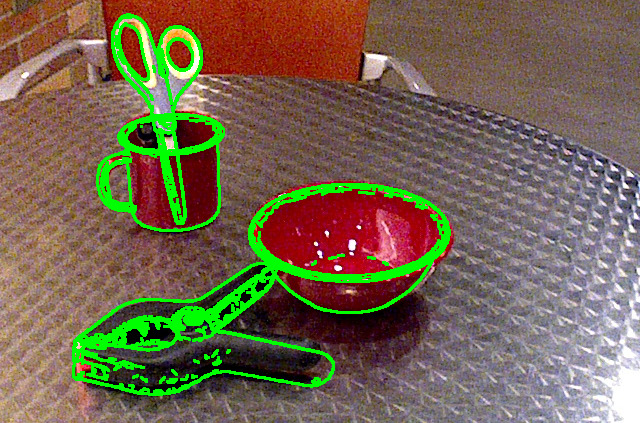}
            };
\node[fill=lightgray, opacity=0.8, text opacity=1, anchor=north east, align=right, inner sep=1pt] at (image.north east) {
                YCB dataset \textbf{\textcolor{green}{end}}
            };
        \end{tikzpicture}
\end{subfigure} \\
    \vspace{0.08cm}
    \begin{subfigure}[t]{1.0\columnwidth}
        \centering
       \begin{tikzpicture}
\node[anchor=south west, inner sep=0] (image) at (0,0) {
                \includegraphics[width=0.48\columnwidth]{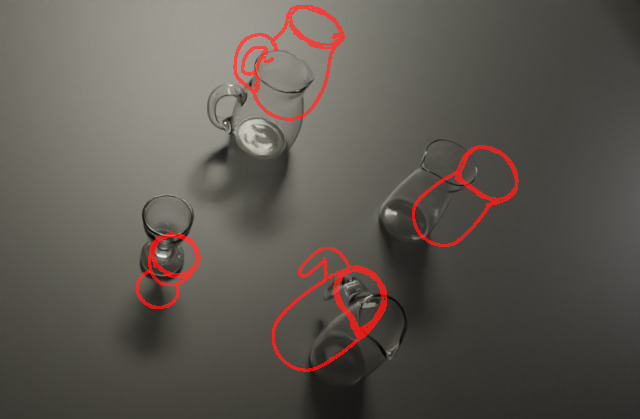}
            };
\node[fill=lightgray, opacity=0.8, text opacity=1, anchor=north east, align=right, inner sep=1pt] at (image.north east) {
                Trans6D-32K dataset \textbf{\textcolor{red}{seed}}
            };
        \end{tikzpicture}
       \begin{tikzpicture}
\node[anchor=south west, inner sep=0] (image) at (0,0) {
                \includegraphics[width=0.48\columnwidth]{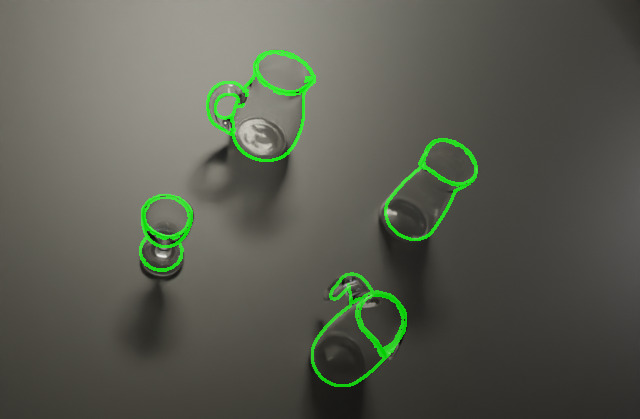}
            };
\node[fill=lightgray, opacity=0.8, text opacity=1, anchor=north east, align=right, inner sep=1pt] at (image.north east) {
                Trans6D-32K dataset \textbf{\textcolor{green}{end}}
            };
        \end{tikzpicture}
\end{subfigure}
    \caption{
        WHS on
        ground sample (top),
        daily~\cite{icar:calli:2015} (middle) and
        transparent objects~\cite{ral:yu:2023} (bottom)
        with $\SI{3}{\centi\meter}$, $\SI{10}{\degree}$ seed error.
    }
    \label{fig:extensionwithtextboxes}
\end{figure}
}

\newcommand{\Mmakeresultstablewithloftr}[1]{
\begin{table*}[#1]
\smallskip
\smallskip
\centering
\begin{tabular}{|l|l|r|r|r|r|r|r|r|r|r|r|r|}
  \cline{3-13}
  \multicolumn{2}{c|}{} & \multicolumn{2}{c|}{Over 2000 runs [\%]} & \multicolumn{3}{c|}{Normal translation [mm]} & \multicolumn{3}{c|}{Lateral translation [mm]} & \multicolumn{3}{c|}{Rotation [mrad]} \\
  \cline{3-13}
  \multicolumn{2}{c|}{} & Completed & Success & Avg. & Std. dev. & Max. & Avg. & Std. dev. & Max. & Avg. & Std. dev. & Max.\\
\hline
\multirow{7}{*}{\rotatebox[origin=c]{90}{\textbf{Rover BC}}} & Input & 100.0 & 0.0 & -0.095 & 12.73 & 19.92 & 16.94 & 8.853 & 29.99 & 48.86 & 26.32 & 86.92 \\
 & ORB & 0.0 & 0.0 & N/A & N/A & N/A & N/A & N/A & N/A & N/A & N/A & N/A \\
 & SSD & 2.1 & 0.9 & 8.179 & 23.76 & 93.76 & 21.97 & 38.31 & 142.2 & 193.3 & 196.8 & 593.4 \\
 & NCC & 94.3 & 94.0 & -0.062 & 1.153 & 23.60 & 0.177 & 1.511 & 57.30 & 2.512 & 15.78 & 330.9 \\
 & LoFTR-R & \textbf{100.0} & 98.0 & -0.174 & 0.324 & 2.009 & 0.156 & 0.153 & 1.332 & 3.936 & 3.782 & 37.95 \\
 & LoFTR-S & 99.5 & 91.0 & -0.205 & 1.421 & 5.842 & 0.318 & 1.792 & 74.83 & 7.232 & 12.24 & 351.3 \\
 & WHS & \textbf{100.0} & \textbf{100.0} & \textbf{0.005} & \textbf{0.082} & \textbf{0.446} & \textbf{0.083} & \textbf{0.036} & \textbf{0.232} & \textbf{1.143} & \textbf{0.708} & \textbf{6.477} \\
\hline
\multirow{7}{*}{\rotatebox[origin=c]{90}{\textbf{Lander OS}}} & Input & 100.0 & 1.1 & -0.054 & 7.321 & 11.45 & 7.906 & 4.131 & 14.00 & 17.13 & 9.225 & 30.47 \\
 & ORB & 45.6 & 21.1 & 0.107 & 2.060 & 14.63 & 0.846 & 0.916 & 7.278 & 25.12 & 25.33 & 148.9 \\
 & SSD & 43.6 & 43.6 & 0.159 & 0.185 & 0.865 & 0.163 & 0.072 & \textbf{0.494} & 2.707 & 1.673 & 11.52 \\
 & NCC & 98.7 & 98.7 & 0.146 & \textbf{0.150} & \textbf{1.053} & \textbf{0.162} & \textbf{0.064} & \textbf{0.494} & \textbf{2.206} & \textbf{1.367} & \textbf{9.351} \\
 & LoFTR-R & 95.0 & 22.3 & 0.059 & 1.699 & 7.662 & 0.917 & 0.907 & 31.99 & 27.21 & 15.11 & 109.3 \\
 & LoFTR-S & 51.1 & 6.6 & -1.103 & 7.535 & 54.16 & 5.085 & 11.69 & 71.67 & 31.81 & 22.02 & 157.4 \\
 & WHS & \textbf{100.0} & \textbf{100.0} & \textbf{0.012} & 0.208 & 1.213 & 0.167 & 0.075 & 0.590 & 2.725 & 1.737 & 13.20 \\
\hline
\end{tabular}
\caption{Quantitative results on synthetic data: run completion and accuracy success rates, and error statistics per dimension.}
\label{table:results_stats_with_loftr}
\end{table*}
}

\newcommand{\Mmakeresultstimingloftr}[1]{
\begin{table}[#1]
\centering
\resizebox{1.0\columnwidth}{!}{
\begin{tabular}{|l|r|r|r|r|r|r|}
  \cline{2-7}
  \multicolumn{1}{l|}{} & ORB & SSD & NCC & LoFTR-R & LoFTR-S & WHS \\
\hline
  Initialization & 1.13 & 1.11 & 1.12 & 1.35 & 1.23 & 1.22 \\
  Rendering      & 7.40 & 7.49 & 7.11 & 8.08 & 7.70 & 6.91 \\
  Matching       & 0.60 & 3.73 & 3.61 & 671.26 & 2146.98 & 13.17 \\
  Pose update    & 2.66 & 1.67 & 0.38 & 0.69 & 4.65 & 0.02 \\
  \hline
  \hline
  \textbf{Total} $[\SI{}{\minute}]$ & 11.79 & 14.00 & 12.22 & 681.38 & 2160.56 & 21.33 \\
\hline
\end{tabular}
}
\caption{Computation times on $\SI{200}{\mega\hertz}$ flight processor.}
\label{table:timing}
\end{table}
}

\newcommand{\Mmakerealresultstable}[1]{
\begin{table*}[#1]
\smallskip
\smallskip
\centering
\begin{tabular}{|l|r|r|r|r|r|r|r|rr|}
  \cline{2-10}
  \multicolumn{1}{r|}{} & \multicolumn{2}{c|}{\textbf{Testbed OS} $20$ runs $[\%]$} & \multicolumn{2}{c|}{\textbf{Testbed BC} $20$ runs $[\%]$} & \multicolumn{2}{c|}{\textbf{Mars BC near} $3$ runs $[\%]$} & \multicolumn{3}{c|}{\textbf{Mars BC far} $3$ runs $[\%]$} \\
  \cline{2-10}
  \multicolumn{1}{r|}{} & Completed & Success & Completed & Success & Completed & Success & Completed & Success & (norm. / lat. / rot.) \\
\hline
  ORB     & 0.0            & 0.0            & 0.0            & 0.0            & 0.0            & 0.0            & 0.0            & 0.0 & (0.0 / 0.0 / 0.0) \\
  SSD     & \textbf{100.0} & 45.0           & 40.0           & 0.0            & 0.0            & 0.0            & 0.0            & 0.0 & (0.0 / 0.0 / 0.0) \\
  NCC     & \textbf{100.0} & 45.0           & 65.0           & 0.0            & 33.3           & 0.0            & 0.0            & 0.0 & (0.0 / 0.0 / 0.0) \\
  LoFTR-R & 70.0           & 5.0            & \textbf{100.0} & 95.0           & \textbf{100.0} & \textbf{100.0} & \textbf{100.0} & 0.0 & (66.7 / \textbf{100.0} / 0.0) \\
  LoFTR-S & 10.0           & 0.0            & \textbf{100.0} & 95.0           & 66.7           & 0.0            & 33.3           & 0.0 & (0.0 / 0.0 / 0.0) \\
  WHS     & \textbf{100.0} & \textbf{100.0} & \textbf{100.0} & \textbf{100.0} & \textbf{100.0} & \textbf{100.0} & \textbf{100.0} & 0.0 & (\textbf{100.0} / \textbf{100.0} / 0.0) \\
\hline
\end{tabular}
\caption{Localization results on real-world data. We decompose success rates per dimension on the far Mars BC.}
\label{table:results_real}
\end{table*}
}

\begin{document}

\maketitle
\begin{abstract}
    We consider the problem of vision-based 6-DoF object pose estimation in the context of the notional Mars Sample Return campaign, in which a robotic arm would need to localize multiple objects of interest for low-clearance pickup and insertion, under severely constrained hardware.
    We propose a novel localization algorithm leveraging a custom renderer together with a new template matching metric tailored to the edge domain to achieve robust pose estimation using only low-fidelity, textureless 3D models as inputs.
    Extensive evaluations on synthetic datasets as well as from physical testbeds on Earth and in situ Mars imagery shows that our method consistently beats the state of the art in compute and memory-constrained localization, both in terms of robustness and accuracy,
    in turn enabling new possibilities for cheap and reliable localization on general-purpose hardware.
\end{abstract}

 \section{Introduction}
\label{sec:introduction}

NASA's \emph{Perseverance} rover, which landed on Mars in 2021, is collecting rock and atmosphere \emph{sample tubes} to be returned to Earth as part of a planned NASA-ESA follow-up program, Mars Sample Return~\cite{aa:muirhead:2020}.
In a potential concept of operations (see Fig.~\ref{fig:conops_and_results}), a \emph{Sample Return Lander} would rendezvous with Perseverance on Mars, use a robotic arm to retrieve sample tubes from the rover's \emph{bit carousel} (BC) and load them into an \emph{orbiting sample} (OS) canister to be launched into orbit then returned to Earth.
The lander arm would initially extend to a ready pose $\SI{3}{\cm}$
in front of the rover BC.
It would then refine its pose knowledge via visual localization using arm-mounted cameras to retrieve a tube from the BC.
The arm, now holding a tube, would then extend to another ready pose $\SI{3}{\cm}$ in front of the OS and
again refine its pose knowledge using vision to enable insertion into the OS.
While lander images could be downlinked for processing on Earth,
such localization would take at least 3 Mars days per tube ($74$ Earth hours) due to limited communication windows,
during which uncertainties could accumulate from thermal expansion and other movement.
In order to fit within a $\SI{30}{\minute}$ time budget, localization must instead be fully autonomous.

\Mmakeconopsandresultsfigure{t}

This is a difficult problem.
First, we must deal with extremely tight requirements due to the cost and criticality of the mission.
For example, the relative pose between the end effector and the BC before localization can only be assumed known within
\SI{75}{\milli\meter} and \SI{5}{\degree} due to kinematics uncertainties, thermal deflections,
sinkage in the Martian sand as the arm extends, etc.
Still, from these large input uncertainties, 
visual localization accuracy is required to be within \SI{0.4}{\milli\meter} and \SI{0.25}{\degree}
in order to guarantee safety of the rover and lander during open-loop tube transfer.
Second, such accuracy must be achieved from only monocular inputs,
allowing for failure of either lander camera.
Third, localization must run on a flight-qualified, radiation-hardened, single-core \SI{200}{\mega\hertz} processor with
only \SI{10}{\mega\byte} of RAM.
While existing methods can tackle some of these challenges, they typically fall short on accuracy and robustness to operational conditions~\cite{iccv:rublee:2011,tpami:goshtasby:1984} or amount of data and compute required~\cite{ral:pham:2021,cvpr:sun:2021}.

In this paper, we propose a novel method that is capable of performing object localization within extremely tight accuracy margins while being robust to noise and lighting, using a single image as input, under severely limited computation and timing budgets.
We do so by building upon the state of the art in multiple areas of visual localization (see Section~\ref{sec:related_work}) and extending it through the following contributions:
\begin{itemize}
    \item A novel render-and-compare algorithm that matches synthetic images from iterative viewpoint hypotheses to a monocular image to localize (see Section~\ref{sec:algorithmoverview});
    \item A custom, extendable renderer focusing on edge features that are more likely to hold between synthetic rendering and reality (see Section~\ref{sec:rasterization});
    \item A novel template matching metric that is explicitly tailored to the edge domain to finish bridging the sim-to-real gap (see Section~\ref{sec:metric});
    \item An extensive dataset comprising 4000 simulated images under the range of experimental conditions we expect to face on Mars, as well as real images collected on Mars and development testbeds on Earth, that we make publicly available\footnote{https://doi.org/10.48577/jpl.9IZDE2} (see Section~\ref{sec:dataset}).
\end{itemize}
We demonstrate the effectiveness of our proposed method by evaluating it against existing works from the state of the art (see Section~\ref{sec:experiments}). We conclude by discussing the challenges we encountered as well as future extensions of our work outside space (see Section~\ref{sec:discussion}),
as we believe cheap, precise and robust localization will become increasingly valuable as more and more reasoning, locomotion and interaction capabilities involving deep learning start competing for resources.

 \section{Related Work}
\label{sec:related_work}

We consider the problem of estimating the 6-DoF pose of an object of known geometry, e.g., defined as a 3D mesh, textured or not.
As surveyed in~\cite{ftcgv:lepetit:2005},
the monocular case has traditionally leveraged different categories of features
such as edges, points and regions.
Edge-based methods~\cite{ijcv:lowe:1992,tip:liu:2024}
have been used in the context of \emph{object tracking}
to refine an object's pose by optimizing the projection of its edges
over consecutive frames.
Region-based methods have demonstrated great results in tracking complex shapes under perturbators such as occlusions by aligning silhouettes with segmented regions of the image~\cite{tpami:tjaden:2018,ijcv:zhong:2019} but still suffer from the need for a pose seed,
and that ultimately the problem is multimodal
(i.e., a given silhouette can be explained by multiple poses).
In contrast, point-based methods have received accrued interest as feature descriptors developed for viewpoint invariance such as Scale-Invariant Feature Transform (SIFT) and Oriented features from accelerated segment test and Rotated Binary robust independent elementary features (ORB)~\cite{ijcv:lowe:2004,iccv:rublee:2011} enabled less constrained matching between arbitrary objects and images.
The pose of an object can then be solved from a set of 2D-3D correspondences using the Perspective-n-Point (PnP) algorithm~\cite{ijcv:lepetit:2009,tpami:xu:2016} and Random Sample Consensus (RANSAC) outlier rejection~\cite{ca:fischler:1981}.

We note that the term \emph{template matching} is overloaded in the context of visual localization
and can refer to the distillation of object characteristics into carefully crafted features for efficient nearest-point retrieval~\cite{tpami:sattler:2016,ijrr:churchill:2013,rss:mcmanus:2014}.
It can also refer to the low-level process of finding the best-matching location of an image patch
within a larger image according to some metric,
e.g., minimizing the sum of squared differences (SSD)~\cite{cvpr:korman:2013,icra:kang:2024} between the template pixels and a sliding window in the full image.
Normalized cross correlation (NCC)~\cite{tpami:goshtasby:1984}
was also used
to localize a rover relative to a high definition model of its surrounding terrain using stereo~\cite{ral:pham:2021}.
Other approaches relying on a render-and-compare framework showed robustness to moderate lighting changes
but still required textured reference models~\cite{icra:pascoe:2015,icra:ok:2016}.
The method we propose builds upon these works to
enable monocular localization
using only low-fidelity meshes as input.

Though not compatible with hardware constraints, approaches using depth from stereo~\cite{iros:stoiber:2023}
could facilitate localization as it is a common challenge for monocular cameras~\cite{cvpr:li:2022}.
Recently, deep learning-based methods were also shown to be able to perform one-shot pose regression~\cite{cvpr:wang:2019,cvpr:peng:2019,cvpr:lin:2024}
or even zero-shot on novel objects by render and compare~\cite{eccv:li:2018,cvpr:ponimatkin:2022}.
Although the applicability of such methods to flight processors is currently hindered
by low compute and memory availability,
as well as strict requirements on interpretability and verifiability,
current efforts on model compression~\cite{tpami:he:2023} could enable on-board inference in the future,
while using focused models as sub-components, e.g., for monocular depth prediction~\cite{tpami:ranftl:2020,cvpr:ke:2024}
or learning-based feature matching~\cite{cvpr:dusmanu:2019,cvpr:sarlin2020,cvpr:sun:2021},
could facilitate verification of end-to-end methods.

 \section{Method}
\label{sec:algorithm}

While prior methods have addressed aspects of this problem, none fully meet our hardware and accuracy constraints.
Therefore, we propose new approaches for image synthesis and template matching that together enable high-accuracy localization on compute and memory restricted environments.

\subsection{Render-and-Compare Algorithm}
\label{sec:algorithmoverview}

In the following, matrix variables are denoted in bold and scalar in italic.
Let $\MTestImage$ denote a \emph{test} image of size $\MTestSize$ captured by the lander arm camera $\MCameraFrame$ in front of the target station $\MStationFrame \in \left\{\text{rover BC}, \text{lander OS}\right\}$.
Let $\MTransform{\MStationFrame}{\MCameraFrame} \in SE(3)$ denote the transformation between the target station and camera frames, which we seek to estimate from vision.
We assume the camera calibrated but do not require the use of a specific model (e.g., pinhole),
only that a mapping between world 3D points and image 2D pixels is available.

First, we populate a virtual scene with our current estimate of the world state, including
a 3D model of the target station.
With $\MCameraEstimateFrame$ the virtual camera frame,
we initialize $\MTransform{\MStationFrame}{\MCameraEstimateFrame}$ at the commanded station ready pose $\MTransform{\MStationFrame}{\MCameraEstimateFrame, 0}$.
Our goal is then is to refine $\MTransform{\MStationFrame}{\MCameraEstimateFrame}$ towards the actual camera pose $\MTransform{\MStationFrame}{\MCameraFrame}$.
Second, we synthesize a \emph{baseline} image $\MBaselineImage$ of the virtual scene as seen from $\MCameraEstimateFrame$,
that we use it as a reference for comparison to the test image $\MTestImage$.
To do so, we write a custom renderer that provides us with an \emph{ideal} edge map as baseline image that we can compare to an \emph{imperfect} edge map derived from the test image.
Third, we look for correspondences between $\MBaselineImage$ and $\MTestImage$ by performing \emph{template matching}.
This consists in choosing a set of $\MTemplateCount$ pixels of interest in the baseline image,
$\MBaselinePixel_{i}$, $i \in [1, \MTemplateCount]$,
extracting a \emph{template}
$\MTemplateImage_{i}$ from $\MBaselineImage$ centered on $\MBaselinePixel_{i}$,
and looking for its best match in $\MTestImage$, $\MTestPixel_{i}$.
At this stage, we have a mapping between 2D baseline and test pixels $\MBaselinePixel_i \xleftrightarrow{} \MTestPixel_i$.
From 3D rendering,
we can also retrieve the depth associated to $\MBaselinePixel_i$
and the corresponding 3D point $\MBaselinePoint_i$ on the station.
As a fourth step, we thus use the resulting 2D-3D correspondences $\MBaselinePoint_i \xleftrightarrow{} \MTestPixel_i$ to calculate the best-fitting camera pose $\MTransform{\MStationFrame}{\MCameraEstimateFrame, 1}$ using PnP-RANSAC~\cite{ijcv:lepetit:2009,ca:fischler:1981}.
We repeat these four steps, iteratively updating the camera pose $\MTransform{\MStationFrame}{\MCameraEstimateFrame, k}$ for up to $n_{\text{max}}$ iterations (see Fig.~\ref{fig:tsm}).

In our experiments, we set $n_{\text{max}} = 10$
and exit early when relative corrections per iteration are within $\SI{0.5}{\milli\meter}$ and $\SI{0.5}{\degree}$ over two consecutive iterations (convergence),
or when absolute pose corrections w.r.t. the seed pose exceed the known uncertainty range from end effector placement (early failure).
We typically observe early convergence after $5$ iterations.
We also set: template sizes at powers of 2 based on the size of potential geometrical features of interest in the image (e.g., OS sleeves, BC screws);
search areas based on the projection of input baseline points composed with worst-case input uncertainties on the commanded pose (e.g., initially translate and rotate by $\pm \SI{30}{\milli\meter}$, $\pm \SI{5}{\degree}$ on each axis, then halve twice over the next two iterations);
and reprojection error thresholds at $\SI{8}{\text{px}}$ initially to maximize confidence in the initial pose estimate then also halve twice.

\Mmakealgorithmfigure{t}

\subsection{Salient Edge Rendering}
\label{sec:rasterization}

We consider ray tracing~\cite{book:glassner:1989} and rasterization~\cite{siggraph:pineda:1988}
to generate an image of the scene from the current camera pose hypothesis.
In the former, the path of individual rays between light source, lens and sensor is simulated,
enabling photorealistic image rendering.
Its computational cost however is incompatible with our flight processor as image rendering could take days.
Instead, we derive our renderer from the latter, which simply consists in iterating over all object polygons
and projecting them into image space.
By keeping track of the lowest depth at each pixel over all polygons,
we obtain a \emph{depth map} $\MDepthBuffer$
and derive an intensity image from the material color associated to those polygons and the view angle.

While existing libraries such as OpenGL~\cite{book:shreiner:2009} can handle most of the rendering logic,
they are usually optimized for pinhole camera models,
whereas real cameras typically include some amount of distortion.
To be able to match renders to real images for localization,
we would need to distort the former \emph{a posteriori}, or undistort the latter,
both resulting in more compute as well as potential loss of information.
As related work also showed that image matching in the edge domain increased robustness to changes in lighting~\cite{ral:pham:2021},
we extend our rasterizer to build \emph{ideal edge maps}.
We do so by now keeping track of the surface normal at each pixel as we build the depth map $\MDepthBuffer$,
then mark as \emph{salient edges} pixels such that surface normals or depths border discontinuities past chosen thresholds.
This lets us calculate likely edges in image space from geometry only,
while completely circumventing potential pitfalls from high-fidelity rendering followed by edge detection
(e.g., computational cost, parameter tuning and scene fidelity).
We set the discontinuity thresholds based on the object's geometry and its discretization as a 3D mesh.
Low thresholds can result in false edges appearing as artifacts from low model resolution,
while high thresholds can result in real edges not being rendered.
As a rule of thumb, we found that $\SI{30}{\degree}$ and $\SI{5}{\milli\meter}$
worked for all our experiments without further tuning,
although individual thresholds could also be applied per object.
In the rest of this paper, we operate in the edge domain and consider as baseline image $\MBaselineImage$ the edge map from salient edge rendering and as test image $\MTestImage$ the edge map from Canny edge detection~\cite{tpami:canny:1986} on the intensity image to localize,
with baseline templates centered on edge pixels for improved robustness.
We found that performing histogram equalization helped Canny edge detection in areas of poor contrast,
at the cost of increased background noise (see Fig.~\ref{fig:tsm}).
In the following, we propose a novel template matching metric that leverages these benefits without the drawbacks by masking this noise out.

\subsection{Weighted Hamming Similarity}
\label{sec:metric}

We consider the task of finding the best match in the $\MTestSize$ test image $\MTestImage$
for a given template image $\MTemplateImage$ extracted as a $\MTemplateSize$ subframe from the baseline image $\MBaselineImage$.
This is typically done by evaluating a similarity score function between the baseline template and a sliding window over the test image,
resulting in a $(\MTestSizeCol - \MTemplateSizeCol + 1) \times (\MTestSizeRow - \MTemplateSizeRow + 1)$ score matrix $\MScoreMatrix$.
Since we operate in the edge domain, test and template images are binary edge maps such that each pixel value is $0$ or $1$.
As our templates are synthetic and we only render the objects to localize without the background,
zero-value pixels can be either no-edge (smooth part of a rendered object) or empty space (unrendered object or background).
Therefore, we incorporate a mechanism to only match the parts of the image that were explicitly rendered and ignore the others.

Let then
$\MMaskImage$ denote a \emph{mask} associated to the baseline template $\MTemplateImage$.
$\MMaskImage$ is a binary mask such that each pixel $\MMaskElement_{u, v}$ is $1$ if the corresponding pixel in the template image $\MTemplateElement_{u, v}$ should be used to evaluate the matching score and $0$ if it should be ignored.
The availability of such masks together with ideal edge maps from our rendering pipeline then lets us
propose a novel, intuitive template matching metric counting pixels that are simultaneously edges or simultaneously non-edges on both template and test image, normalized by the number of unmasked edge $\MTemplatePixelCountEdge$ and non-edge $\MTemplatePixelCountNotEdge$ pixels in the template image.
We call this new metric Weighted Hamming Similarity (WHS) and define its score function as:
\begin{align}
    \label{eq:matching_score_full}
    & \MScoreElement_{i, j} = \MTemplateWeightEdge \MScoreElement_{i, j}^{\MEdgeMarker} + \MTemplateWeightNotEdge \MScoreElement_{i, j}^{\MNotEdgeMarker},
    \text{ with:} \\
    \label{eq:matching_score_weight_edge}
    & \MTemplateWeightEdge = 
\frac{1}{\MTemplatePixelCountEdge} \text{ if $\MTemplatePixelCountEdge > 0$ and }
\MTemplateWeightNotEdge =
\frac{1}{\MTemplatePixelCountNotEdge} \text{ if $\MTemplatePixelCountNotEdge > 0$, else } 0, \\
& \MScoreElement_{i, j}^{\MEdgeMarker} = 
    \sum\limits_{u = 0}^{\MTemplateSizeRow - 1} \sum\limits_{v = 0}^{\MTemplateSizeCol - 1}
    \MMaskElement_{u, v} \cdot (\MTemplateElement_{u, v} == \MTestElement_{i+u, j+v} == 1), \\
    \label{eq:matching_score_not_edge_full}
    & \MScoreElement_{i, j}^{\MNotEdgeMarker} = 
    \sum\limits_{u = 0}^{\MTemplateSizeRow - 1} \sum\limits_{v = 0}^{\MTemplateSizeCol - 1}
    \MMaskElement_{u, v} \cdot (\MTemplateElement_{u, v} == \MTestElement_{i+u, j+v} == 0).
\end{align}
Since elements of $\MTemplateImage$ and $\MTestImage$ are $0$ and $1$, this is equivalent to:
\begin{align}
    \label{eq:matching_score_edge_rewrite}
    & \MScoreElement_{i, j}^{\MEdgeMarker} = 
    \sum\limits_{u = 0}^{\MTemplateSizeRow - 1} \sum\limits_{v = 0}^{\MTemplateSizeCol - 1}
    \MMaskElement_{u, v} \cdot \MTemplateElement_{u, v} \cdot \MTestElement_{i+u, j+v}, \\
\label{eq:matching_score_not_edge_rewrite}
    & \MScoreElement_{i, j}^{\MNotEdgeMarker} = 
    \sum\limits_{u = 0}^{\MTemplateSizeRow - 1} \sum\limits_{v = 0}^{\MTemplateSizeCol - 1}
    \MMaskElement_{u, v} \cdot (1 - \MTemplateElement_{u, v}) \cdot (1 - \MTestElement_{i+u, j+v}).
\end{align}

We introduce $\MTemplateMaskedEdgeImage$ (resp. $\MTemplateMaskedNotEdgeImage$) the binary matrix of masked-in edge (resp. not edge) template pixels
and $\MTestEdgeImage$ (resp. $\MTestNotEdgeImage$) the binary matrix of edge (resp. not edge) test pixels.
With $\odot$ the element-wise multiplication and $\mathbf{1}$ the all-one matrix:
\begin{align}
\MTemplateMaskedEdgeImage & = \MMaskImage \odot \MTemplateImage,
     & \MTemplateMaskedNotEdgeImage & = \MMaskImage \odot (\mathbf{1} - \MTemplateImage), \\
\MTestEdgeImage & = \MTestImage,
     & \MTestNotEdgeImage & = \mathbf{1} - \MTestImage,
\end{align}
the full score matrix $\MScoreMatrix$ can finally be written as the weighted sum of two convolutions with reversed kernels $\MTemplateMaskedEdgeImageReversed$ and $\MTemplateMaskedNotEdgeImageReversed$:
\begin{align}
    \label{eq:whs_full}
    \MScoreMatrix = \MTemplateWeightEdge \cdot (\MTemplateMaskedEdgeImageReversed \circledast \MTestEdgeImage) +
    \MTemplateWeightNotEdge \cdot (\MTemplateMaskedNotEdgeImageReversed \circledast \MTestNotEdgeImage).
\end{align}
Importantly, Fast Fourier Transforms (FFT)~\cite{book:nussbaumer:1982} let us evaluate these convolutions efficiently in the frequency domain.

 \section{Datasets}
\label{sec:dataset}

\Mmakedatasetfigure{t}

One challenge specific to our study is the scarcity of true \emph{in situ} data from Mars.
We thus supplement it with testbeds on Earth as well as extensive software simulation (see Fig.~\ref{fig:dataset_full}).

\paragraph{In Situ Mars Imagery}

In between science activities on Mars, we were able to request the collection of $6$ images of the rover BC as observed by its end effector camera~\cite{ssr:maki:2020},
including $3$ times of the day (morning, noon, afternoon) and $2$ camera standoffs:
\emph{near} ($\SI{0.49}{\meter}$) and \emph{far} ($\SI{1.23}{\meter}$).
Pose seeds for localization were derived from rover encoders and forward kinematics
while ground truth for error estimation was calculated from manual annotation and PnP.

\paragraph{Development Testbeds}

Separately, we constructed a two-camera assembly representative of the current lander arm design
and collected $20$ images of an engineering model of the rover BC
as well as $20$ images of a 3D print of the current lander OS design,
with the scene purposely set to challenge our algorithm with lighting, lens and shadowing artifacts.
Ground truthing was again performed by manual annotation, solving over both rigidly-linked cameras using~\cite{eccv:wald:2020}.

\paragraph{Synthetic Data}

Using the Blender~\cite{misc:blender:2024} graphics engine, we build a virtual scene simulating the entire sample tube transfer pipeline, including ground and atmosphere effects, as well as 3D models for the entire spacecraft.
We then explore the range of possible operational conditions by randomly sampling
sun angle, lander and rover poses, and errors on end effector placement w.r.t. the commanded ready pose.
On the OS, we also randomize which sleeve we are inserting into and which sleeves have already been filled, as well as the pre-placed tube clocking angles.
We sample $1000$ combinations of such experimental conditions
and generate physically realistic test images by ray-tracing for each station and camera,
totaling $4000$ images annotated with perfect ground truth from posing in simulation.

\section{Experiments}
\label{sec:experiments}

We now conduct an extensive evaluation of our algorithm performance (accuracy and computation time) on synthetic data as well as real Earth and Mars imagery (see Fig.~\ref{fig:conops_and_results}).

\subsection{Performance Metrics}

\Mmakeresultsfigurewithloftr{}

For a given test image,
we compose the camera pose w.r.t. the station from visual localization $\MTransform{\MStationFrame}{\MCameraEstimateFrame}$
with the ground truth $\MTransform{\MStationFrame}{\MCameraFrame}$
to calculate the camera pose error as
$\MTransform{\MCameraFrame}{\MCameraEstimateFrame} = \MTransform{\MStationFrame}{\MCameraFrame}^{-1} \cdot \MTransform{\MStationFrame}{\MCameraEstimateFrame}$.
Since error budgets are defined at the end effector level for sample transfer,
we compose $\MTransform{\MCameraFrame}{\MCameraEstimateFrame}$ with
$\MTransform{\MCameraFrame}{\MEndEffectorFrame}$ the transformation between camera and end effector,
which is fixed from mechanical design, therefore also
$\MTransform{\MCameraEstimateFrame}{\MEndEffectorEstimateFrame} = \MTransform{\MCameraFrame}{\MEndEffectorFrame}$.
We thus define the end effector localization error as
$\MTransform{\MEndEffectorFrame}{\MEndEffectorEstimateFrame} \coloneqq
\MTransform{\MCameraFrame}{\MEndEffectorFrame}^{-1} \cdot \MTransform{\MCameraFrame}{\MCameraEstimateFrame} \cdot \MTransform{\MCameraEstimateFrame}{\MEndEffectorEstimateFrame}$.

For each run, we evaluate the localization accuracy by decomposing the 6-DoF error into three elements of interest:
normal and tangential translation errors,
respectively obtained by projecting the translation component of $\MTransform{\MCameraFrame}{\MCameraEstimateFrame}$ 
along the camera depth axis and the image plane,
and out-of-plane rotation errors,
measured as the tilt between the estimated and ground-truth camera depth axis.
Since both the sample tube to pickup and the sleeve to insert into are rotationally symmetric, in-plane rotation is not a concern,
though all our WHS results demonstrated robustness to sampled rotations within a $\SI{10}{\degree}$ uncertainty window from kinematics.
We consider a localization run as \emph{completed} if it converges to a pose at all (even if inaccurate),
and as \emph{successful} if all three error components are within requirements across all dimensions simultaneously.
As baselines to our algorithm, we also evaluate alternative methods fitting compute, memory and verifiability constraints pertaining to this task described in Section~\ref{sec:related_work}:
two FFT-accelerated template matching metrics --
NCC and SSD --
and feature matching using ORB (up to 400 features, 8 pyramids).
For these baselines, we use public implementations from the OpenCV library~\cite{book:bradski:2008}.
For the sake of completeness, we also evaluate the deep-learning-based Local Feature TRansformer (LoFTR) model~\cite{cvpr:sun:2021}, with pre-trained weights provided by the authors.
While it does not fit within mission constraints (\SI{3.8}{\giga\byte} of RAM used at inference time, \SI{10}{\mega\byte} available on the flight computer), its ability to perform local feature matching without retraining makes it a good candidate for safety-critical systems since correspondences can still be validated through PnP-RANSAC.
We evaluate two variants of the LoFTR method on our dataset using the largest model we could fit on our test computer, operating on $1024 \times 1024$ pixel images rather than $2048 \times 2048$ full size.
In the first variant (LoFTR-R), we run our algorithm on test images \emph{resized} to $1024 \times 1024$ at every iteration.
In the second variant (LoFTR-S), we only resize test images at the first iteration in order to calculate the initial camera pose correction, then do the matching at the full resolution split into four $1024 \times 1024$ \emph{subframes} for subsequent iterations.

\subsection{Synthetic Evaluation}

\Mmakeresultstablewithloftr{t}

We evaluate each algorithm against $2000$ images for each station
and depict the resulting error distributions in Fig.~\ref{fig:results_sim_with_loftr}, starting with
\emph{input errors} before localization in blue, measured as the difference before the commanded ready pose and the actual achieved pose.
Configurations that result in a \emph{failed localization} default back to the ready pose as best estimate and therefore keep their initial input errors, in red.
Completed runs are plotted in green, with dotted lines representing accuracy requirements per dimension.
We report detailed statistics in Table~\ref{table:results_stats_with_loftr}
along with sample results in Fig.~\ref{fig:conops_and_results}.

First, we observe that ORB failed on all rover BC test cases as well as most lander OS test cases.
We attribute this to the large discrepancy between physically realistic ray-traced test images and low fidelity rasterization making low-level descriptor matching difficult.
The problem seems exacerbated on the BC as its geometry contains fewer corners than the OS.
Still, even then, less than half of the $45.6 \%$ completed localizations are within requirements.
We observe similar results for the SSD template matching metric.
This is not surprising either due to its inherent sensitivity to absolute intensity values rather than relative distributions, which basic rendering techniques cannot capture.

As NCC is more robust to such illumination changes, it also achieved close to perfect completion rates, with $94.3 \%$ on the rover BC and $98.7 \%$ on the lander OS.
While on the lander OS, all completed runs also converged within requirements and with the tightest error distributions across all methods,
a few of the rover BC runs converged outside requirements.
We refer to those as \emph{false positives}.
Since the end effector is only known to be within $\pm \SI{30}{\milli\meter}$ from the commanded pose,
the error between the true pose and a wrong estimate can be up to $\SI{60}{\milli\meter}$.
One such false positive for NCC on the rover BC resulted in a $\SI{57.3}{\milli\meter}$ lateral error,
which is a \emph{critical failure mode} from a mission safety perspective.
While it could be mitigated by adding redundancies to the localization process (e.g., combining localization from multiple viewpoints),
resulting delays and costs would have to be weighed against other mission requirements.

Finally, we investigate the potential of dense feature matching using deep learning.
On the rover BC, the resize variant (LoFTR-R) achieves even better completion and success rates than NCC (resp. $100 \%$ and $98 \%$)
but still produces $2 \%$ of false positives exceeding rotational requirements.
Perhaps surprisingly, the full resolution subframe variant (LoFTR-S) performs slightly worse than LoFTR-R, with $99.5 \%$ completion and $91 \%$ success rates
(we ensure that LoFTR-R and LoFTR-S first iterations coincide by using the same random seed in our test runs).
Failure inspection suggests that the model attempts to search for keypoints even in largely textureless areas of the image,
which results in noisier correspondences at full resolution since those areas then occupy $4 \times$ the pixels.
On the lander OS, this performance difference is compounded by the fact that LoFTR-R now converges $95 \%$ of the time, but only $22.3 \%$ within requirements.
Failure inspection shows that repeating patterns in the lander OS geometry
cause LoFTR-R to erroneously match different sleeves with each other,
resulting in pose estimates off by one or multiple sleeve's widths.

In contrast, WHS achieved $100 \%$ success rate on both stations.
While it produced slightly wider error distributions than NCC on the lander OS,
it also proposed no false positive.

\subsection{Real-World Evaluation}

\Mmakerealresultstable{t}

We now evaluate all methods on real images from testbeds on Earth
as well as the current rover on Mars.
For the sake of conciseness, we report completion and success rates directly in Table~\ref{table:results_real}.
ORB and SSD show similar trends as on the synthetic dataset, with complete failure for the former and less than $50 \%$ success for the latter on the testbed OS.
However, while NCC previously achieved $98.7 \%$ and $94.0 \%$ success rates on the synthetic OS and BC, respectively,
these fall to $45 \%$ and $0 \%$ on their testbed and Mars variants.
This illustrates that while NCC could successfully handle appearance changes resulting from different rendering methods, the gap between basic onboard shading and reality induces additional challenges.
While this could theoretically be mitigated by increasing the fidelity of onboard rendering,
even ignoring timeline requirements,
it is unclear how precisely the environment can be characterized for physically-realistic rendering
(e.g., precise position of the sun, atmospheric density, dust deposit on rover parts).
In contrast, WHS readily demonstrates robustness to low-fidelity rendering,
with $100 \%$ success rate on all testbed scenarios and the near rover BC despite years of unmodeled dust accumulation
and even worse surface resolution ($\SI{166}{\micro\meter\per px}$ for the rover camera at $\SI{49}{\centi\meter}$ vs. $\SI{128}{\micro\meter\per px}$ for the planned lander cameras).
We note that LoFTR-R also achieved great accuracy on both testbed and Mars BC,
with $95 \%$ and $100 \%$ success rates,
though only $5 \%$ on the testbed OS again due to confusion between repeating geometry.
This illustrates the potential for deep learning methods to bridge the gap between simulation
and reality provided sufficient compute.

Finally, while no method was able to meet all requirements at the far standoff,
further inspection per dimension shows that WHS still managed to meet requirements on normal and lateral translation but not rotation (LoFTR-R also failed lateral).
This is reasonable since a $\SI{123}{\centi\meter}$ distance for the rover end camera now corresponds to a
$\SI{417}{\micro\meter\per px}$ surface resolution
($3 \times$ worse than for the planned lander cameras).
At this distance and resolution, given the rover BC $\SI{15}{\centi\meter}$ radius,
a $\SI{16}{\milli\radian}$ requirement-breaking rotation
would result in image features moving by less than $1/20$-th of a pixel, making it essentially imperceptible.
Overall, these results on physical testbed and in situ Mars data illustrate that
our method combining salient edge rendering with WHS can readily
bridge the gap between low-fidelity rendering and reality under degraded observation conditions,
though ultimately limited by surface resolution.

\subsection{Computation Times}
\label{sec:computationtimes}

\Mmakeresultstimingloftr{t}

We break down timings per method as follows:
1) initialization (image normalization, mesh loading, etc.);
2) viewpoint hypothesis rendering (intensity, edge, depth);
3) feature (ORB, LoFTR) or template matching (SSD, NCC, WHS);
and 4) pose update (PnP-RANSAC).
We report average timings summed up over rover BC and lander OS on a single-core $\SI{200}{\mega\hertz}$ processor in Table~\ref{table:timing}.
As LoFTR cannot run on the flight computer ($\SI{3.8}{\giga\byte}$ required vs. $\SI{10}{\mega\byte}$ available)
and to make runtimes tractable,
we extrapolate timings from single-threaded runs on a $\SI{4.5}{\giga\hertz}$ AMD Ryzen Threadripper 3960X CPU.

We observe the following.
First, timings for steps 1) and 2) are in family across all methods.
This is expected since all use the same 3D models and initial viewpoints.
As the rendering step always takes a relatively large portion of the time budget (about \SI{7}{\min} out of \SI{30}{}), we identify it as a potential target for future optimization.
Step 3) is the main differentiator.
ORB feature matching is extremely fast (less than \SI{1}{\minute}), as both the detection and description steps are computationally cheap, as well as the subsequent distance minimization step.
SSD and NCC are about $6$ times slower as convolutions between initially large templates and test images are costlier.
WHS is another $4$ times slower in our current implementation,
though we believe it could be brought down to a theoretical $2\times$
by using optimized FFT and data structures,
since the only computation difference is that we calculate double the convolutions per Eq.~(\ref{eq:whs_full}).
Further improvements include calculating reusable FFTs from larger test image subframes that can be shared by multiple baseline templates
(as opposed to calculating a new each optimal subframe and FFT for each),
attempting PnP-RANSAC as correspondences are collected and possibly exiting early instead of matching all possible templates at once,
reusing or rejecting input templates based on their performance in past iterations,
and running the first iteration at lower resolution since it typically accounts, due to higher initial uncertainty, for $50 \%$
of the total runtime over $10$ iterations.
On the opposite end of the spectrum, LoFTR-R and LoFTR-S would respectively take about $\SI{11}{}$ and $\SI{36}{\hour}$.
While timing extrapolation does not account for differences in CPU architectures, memory usage and other interactions,
these coarse results still indicate that deep models would require faster radiation-hardened compute units
or different architectures (GPU, FPGA) to be viable in space.
In conclusion, while the WHS pipeline is about two times slower than NCC overall, it remains within timing budgets, leaving comfortable margins over $40 \%$ while also producing greater-quality matches,
which is apparent from PnP-RANSAC converging in fewer iterations in step 4).

 \section{Discussion}
\label{sec:discussion}

\Mmakeextensionfigurewithtextboxes{t}

In this paper, we presented a novel monocular object localization pipeline capable of localizing objects within extremely tight margins
on compute-constrained platforms
-- a topic that could conversely unlock new possibilities for robust and cheap localization on general-purpose hardware.
Our method, leveraging salient edge rendering and a novel template matching metric,
achieved $100 \%$ localization success rate within time allocations and demonstrated robustness to domain shifts from low to high fidelity rendering, images from physical testbeds, and even in situ Mars imagery,
never proposing critical failure modes.
We note that template matching methods generally remain sensitive to in-plane rotation errors.
We depict in Fig.~\ref{fig:extensionwithtextboxes} a failure case where we attempt to localize a sample tube on the ground (a contingency scenario that would involve a different lander architecture~\cite{pr:posada:2025})
and the pose seed is rotationally closer to the cast shadow than the actual tube.
As a workaround, we found that we could start independent localization runs
from pose seeds separated by $\SI{20}{\degree}$ increments
and proceed with the highest-inlier pose after the first iteration.
As future work, we plan to improve robustness to large input errors by integrating template warping in the search process~\cite{isvc:hofhauser:2008,cvpr:korman:2013},
optimizing poses over multiple viewpoints~\cite{eccv:wald:2020,ral:pham:2021},
and evaluating the integration of smaller learning-based components such as
monocular depth prediction~\cite{cvpr:ke:2024} for improved pose validation.
We also ran our method on public datasets and
achieved successful localization even on the challenging case of transparent objects without any algorithm change or parameter tuning.
Though currently running at $1$ frame per second on a $\SI{2.4}{\giga\hertz}$ CPU,
we believe future optimizations
using modern frameworks to get closer to the theoretical $2\text{x}$ speed compared to NCC
and parallelized template search and rendering could enable cheap and robust localization at interactive framerates
for general-purpose robots in daily environments.

 {
\bibliographystyle{IEEEtran}
    \bibliography{bib/paper}
}

\end{document}